\documentclass[10pt,twocolumn,letterpaper]{article}

\usepackage{cvpr}              
\usepackage{lineno}
\usepackage{multirow}
\usepackage{booktabs}
\usepackage{colortbl}
\usepackage{makecell}
\usepackage{lipsum}
\usepackage{mathtools}

\usepackage{xcolor}
\definecolor{V}{RGB}{21,137,139}
\definecolor{X}{RGB}{234,120,60}

\definecolor{cvprblue}{rgb}{0.21,0.49,0.74}
\usepackage[pagebackref,breaklinks,colorlinks,allcolors=cvprblue]{hyperref}

\def\paperID{1014} 
\def\confName{CVPR}
\def\confYear{2025}

\title{DPC: Dual-Prompt Collaboration for Tuning Vision-Language Models}

\author{
    Haoyang Li$^{1,2}$ \quad
    Liang Wang$^{1,2}$ \quad 
    Chao Wang$^{1}$\thanks{Corresponding authors.} \quad
    Jing Jiang$^{2}$ \quad
    Yan Peng$^{1}$\footnotemark[1] \quad
    Guodong Long$^{2}$\footnotemark[1]  \\
    {$^{1}$Shanghai University}, {$^{2}$University of Technology Sydney} \\
    {\tt\small haoyang.li-3@student.uts.edu.au, \{cwang, pengyan\}@shu.edu.cn, guodong.long@uts.edu.au}
}

\begin{document}
\maketitle
\begin{abstract}
The Base-New Trade-off (BNT) problem universally exists during the optimization of CLIP-based prompt tuning, where continuous fine-tuning on base (target) classes leads to a simultaneous decrease of generalization ability on new (unseen) classes. Existing approaches attempt to regulate the prompt tuning process to balance BNT by appending constraints. However, imposed on the same target prompt, these constraints fail to fully avert the mutual exclusivity between the optimization directions for base and new. As a novel solution to this challenge, we propose the plug-and-play \textbf{D}ual-\textbf{P}rompt \textbf{C}ollaboration (\texttt{DPC}) framework, the first that decoupling the optimization processes of base and new tasks at the \textbf{prompt} level. Specifically, we clone a learnable parallel prompt based on the backbone prompt, and introduce a variable Weighting-Decoupling framework to independently control the optimization directions of dual prompts specific to base or new tasks, thus avoiding the conflict in generalization. Meanwhile, we propose a Dynamic Hard Negative Optimizer, utilizing dual prompts to construct a more challenging optimization task on base classes for enhancement. For interpretability, we prove the feature channel invariance of the prompt vector during the optimization process, providing theoretical support for the Weighting-Decoupling of \texttt{DPC}. Extensive experiments on multiple backbones demonstrate that \texttt{DPC} can significantly improve base performance without introducing any external knowledge beyond the base classes, while maintaining generalization to new classes. Code is available at: \href{https://github.com/JREion/DPC}{https://github.com/JREion/DPC}.

\end{abstract}  

\section{Introduction}

\begin{figure}[h]
  \centering
  \includegraphics[width=\linewidth]{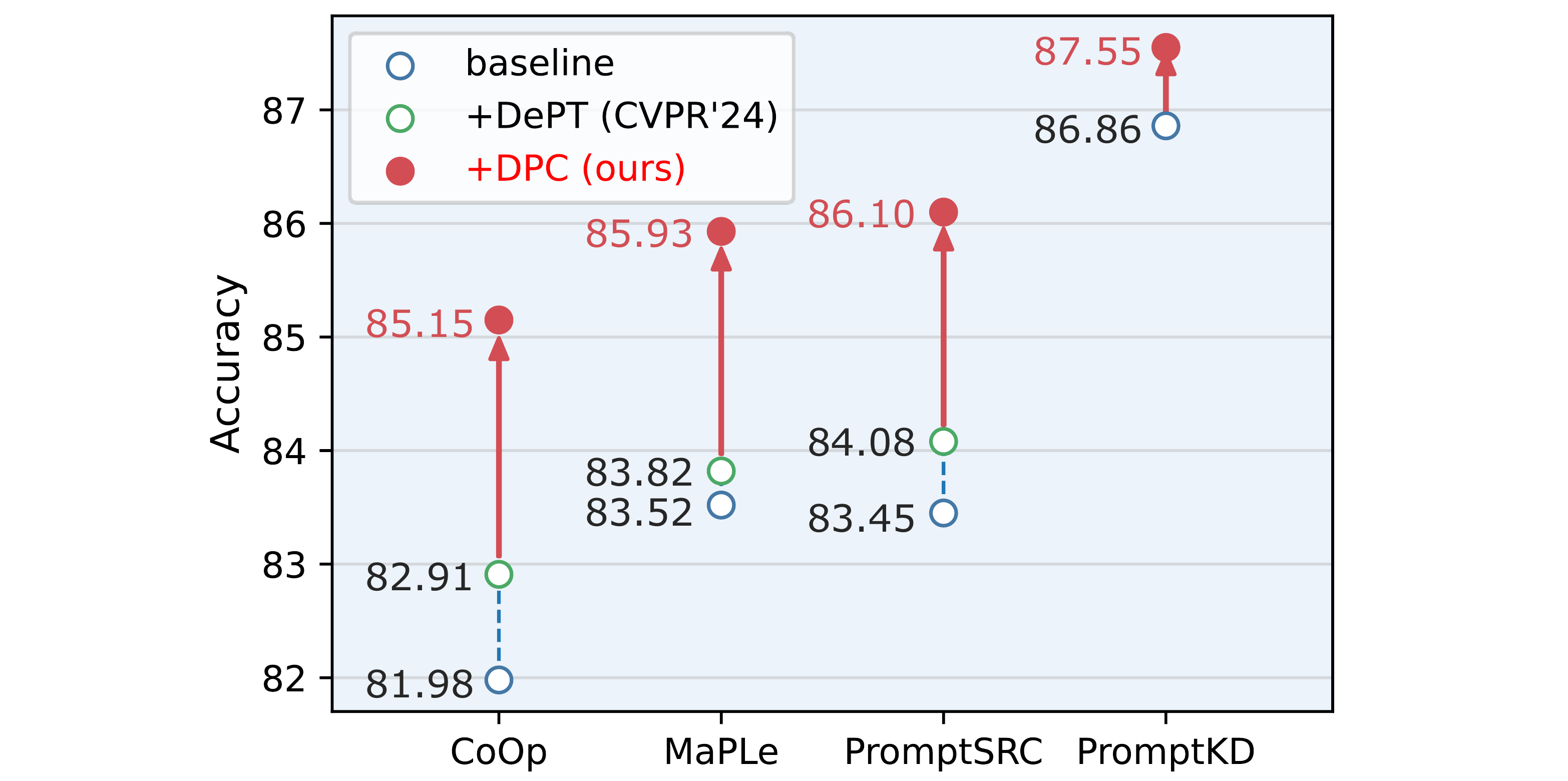}
  \caption{Average classification accuracy of 4 mainstream prompt tuning backbone models on base (target) classes over 11 datasets. \texttt{DPC} achieves state-of-the-art performance compared with baselines and another leading plug-and-play model DePT \cite{zhang2024dept}. }
  \label{Figure 1}
\end{figure}

Vision-Language Models (VLMs), represented by CLIP \cite{radford2021clip}, have revealed cogent cross-modal open-domain representation and zero-shot capabilities. To further efficiently utilize pre-trained VLMs, Prompt Tuning acquires significant attention as a Parameter-Efficient Fine-Tuning (PEFT) method \cite{zhou2022coop,zhou2022cocoop}. Freezing the vision and text encoders, it employs a learnable lightweight prompt vector as a query to guide the output of CLIP towards the target task.

Unfortunately, the optimization process of prompt tuning often encounters a Base-New Trade-off (BNT) problem \cite{zhou2022cocoop,zhang2024dept}. As the prompt increasingly aligns with the target task, the model may overfit to the base (target) classes, resulting in reduced generalization performance on new (unseen) classes. To alleviate the BNT problem, previous approaches attempt to adjust loss functions \cite{yao2023kgcoop, khattak2023promptsrc}, apply constraints to prompts \cite{zhou2022cocoop, yao2024tcp}, add extra feature extractors \cite{gao2024clipadapter,roy2023coprompt}, or involve external knowledge  \cite{zhang2023cafo,li2024promptkd, khattak2024protext}. However, all these methods treat prompts as a single entity to be optimized for a balanced performance between base and new classes. Due to the shift of data distribution, the optimization directions of the two are likely to interfere with each other, making it tough to achieve the global optimum.

\begin{figure}[h]
  \centering
  \includegraphics[width=\linewidth]{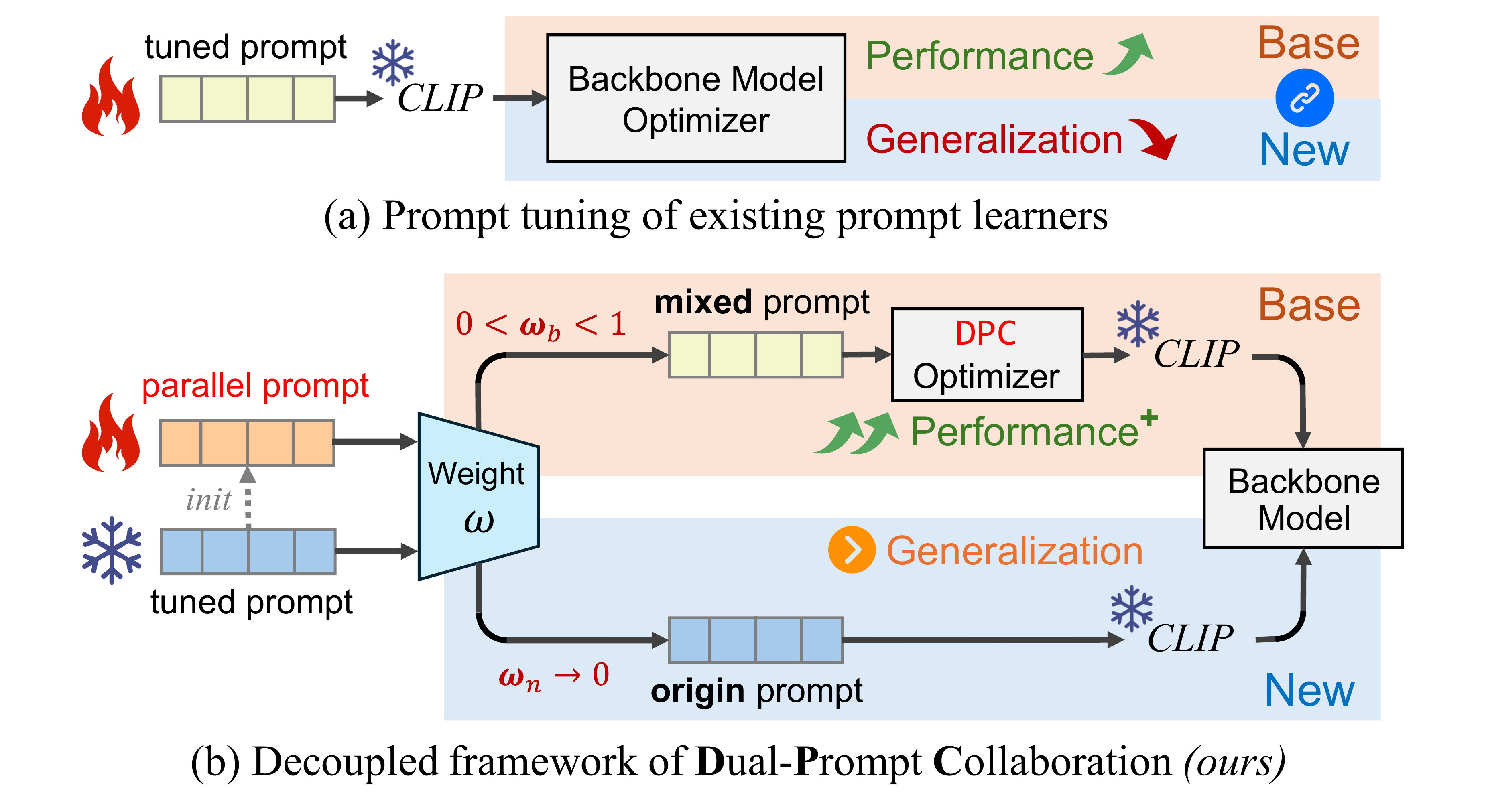}
  \caption{Architecture comparison between (a) existing prompt learners that encounter Base-New Trade-off (BNT) problem and (b) our Dual-Prompt Collaboration framework that decouples the optimization directions of base and new tasks at \textbf{prompt} level.}
  \label{Figure 2}
\end{figure}

To mitigate such interference caused by the mutual exclusivity of optimization directions in prompt tuning for base or new tasks, we propose the Dual-Prompt Collaboration (\texttt{DPC}) framework. This approach is the first to introduce dual prompts optimized in two distinct directions, overcoming BNT problem by decoupling base and new tasks at the \textbf{prompt} level. Since the optimization in prompt tuning basically targets the learnable prompts, we believe that prompt-level decoupling is more fundamental. Specifically, based on the tuned prompt vector obtained by backbone prompt learner (e.g., CoOp \cite{zhou2022coop}), we initialize and activate a separate parallel prompt. The two prompts are independently utilized for new class generalization and base class enhancement, respectively. To quantitatively regulate the optimization directions of dual prompts, we establish a flexible weight adjustment framework named \emph{Weighting-Decoupling}. This module introduces a task-specific alterable hyperparameter $\omega$, allowing dynamic adjustment of the weight distribution between dual prompts, thus preventing overfitting during base class optimization while fully retaining the generalization ability for unseen classes of the backbone model (typically when $\omega\to0$). This framework fixes BNT problem by better controlling fine-tuning directions.

For the interpretability of this structure, in Section \ref{Sec4.4}, we prove the invariance of feature channels in the prompt vectors during fine-tuning, i.e., continuous prompt optimization of \texttt{DPC} does not compromise the feature distribution of the prompt vectors. This indicates that the prompt tuning process can be correctly measured by \texttt{DPC} weights without causing feature bias.

Meanwhile, to further strengthen base class performance, we devise a \emph{Dynamic Hard Negative Optimizer} to fine-tune the parallel prompt. This module is used to construct and learn to distinguish hard negative samples to further match the base classes. We first reuses the prompt tuning backbone with the collaboration of tuned prompt to spontaneously obtain Top-$K$ negative results on the base classes as hard negative objects. Next, a Feature Filtering module applying L2 normalization is appended to extract hard negative text features aligned with paired images, while maintaining the distribution of base classes. Subsequently, we introduce an improved symmetric cross-entropy loss as an additional optimization term, constructing a more challenging vision-language contrastive learning task. This approach facilitates the \texttt{DPC} to more deeply fit the latent feature distribution of the base classes, while enhancing feature alignment between the vision and language modalities.

Our \texttt{DPC} is orthogonal to most existing prompt tuning backbones, exhibiting outstanding plug-and-play characteristics. Additionally, our model is self-contained, requiring no external knowledge beyond the train splits of base classes. In experimental part, we apply \texttt{DPC} in 4 backbone models \cite{zhou2022coop, khattak2023maple, khattak2023promptsrc, li2024promptkd} with different forms of learnable prompts and conduct base-to-new generalization tasks on 11 recognition datasets. Results in \cref{Figure 1} denote that \texttt{DPC} significantly enhances the base class performance in most of backbone models and datasets, while non-destructively preserves the generalization capability of the backbones.

Our main contributions can be generalized as follows:

\begin{enumerate}[1)]
    \item We propose Dual-Prompt Collaboration (\texttt{DPC}) with flexible Weighting-Decoupling structure. To the best of our knowledge, this is the first prompt tuning enhancement strategy that decouples at the prompt level to overcome the BNT problem. 
    \item We design a novel Dynamic Hard Negative Optimizer, significantly enhancing the base class performance of \texttt{DPC} by establishing harder visual-text aligning tasks using dual prompts, achieving new State-Of-The-Art.
    \item We introduce plug-and-play and self-contained features to the model, endowing it with outstanding adaptability and transferability while minimizing requirements of external knowledge.
\end{enumerate}

\begin{figure*}[t]
  \centering
  \includegraphics[width=\textwidth]{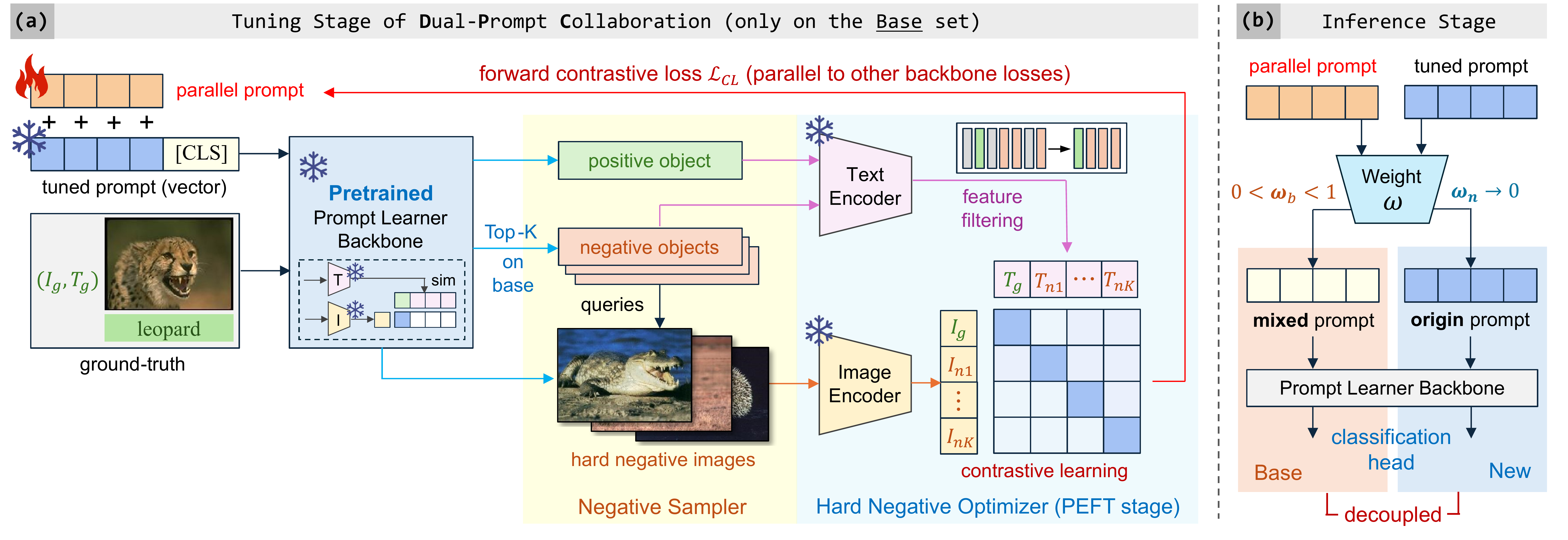}
  \caption{Overview of our proposed \texttt{DPC}. In (a) fine-tuning stage, \texttt{DPC} initializes parallel prompt $\boldsymbol{P}^{\prime}$ based on tuned prompt $\boldsymbol{P}$ obtained by fine-tuning backbone. Negative Sampler applies tuned prompt $\boldsymbol{P}$ as query to sample hard negatives, then feed them into HNO optimizer to enhance base tasks. In (b) inference stage, \texttt{DPC} decouples base and new tasks by independent weight accumulation on dual prompts. }
  \label{Figure 3}
\end{figure*}

\section{Related Work}

\noindent \textbf{Prompt Tuning in VLMs.} As frameworks that deeply integrate and align visual and textual modalities, Vision-Language Models (VLMs) \cite{lu2019vilbert, radford2021clip, jia2021align, li2023blip} have recently gained extensive attention, demonstrating remarkable potential in multiple cross-modal reasoning tasks. However, with the expansion of VLM network parameters, full-parameter fine-tuning requires substantial computational costs. In contrast, prompt tuning is proposed as a Parameter-Efficient Fine-Tuning (PEFT) technique on frozen pre-trained VLMs \cite{han2024peft}. Entirely different from the template-based prompts utilized in CLIP or Large Language Models (LLMs) \cite{li2021prefix, wei2022chain}, prompt tuning introduces a set of learnable lightweight vectors for replacing manually constructed hard prompts, and preserves the learnability of only prompt vectors during fine-tuning. As queries, these prompt vectors are continuously optimized, guiding the outputs of VLMs towards domain-specific data. Numerous approaches propose various forms of learnable prompts for constructing text features \cite{zhou2022coop, zhu2023prograd, tian2024argue}, image features \cite{jia2022vpt, pei2024sa2vp, xu2024provp}, or joint visual-textual encoding \cite{zang2022upt, khattak2023promptsrc, zhang2024cocole}. Although the design of prompts varies across models, in our work, \texttt{DPC} constructs parallel prompts following the structure of backbone models, regardless of prompt forms.

\noindent \textbf{Base-New Trade-off of Prompt Tuning.} The Base-New Trade-off (BNT) problem of CLIP-based prompt learner is first put forward in CoCoOp \cite{zhou2022cocoop}. Essentially, it is due to the overfitting of prompt vector to the distribution of base classes after optimization, thereby reducing generalization to new classes. Numerous efforts are made to mitigate the impact of the BNT problem. Some approaches focus on appending generalization-related constraints during the prompt optimization process, like conditional context \cite{zhou2022cocoop}, semantic distance balancing \cite{yao2023kgcoop}, or consistency loss \cite{khattak2023promptsrc}. Other studies introduce additional feature extractors like Adapters \cite{gao2024clipadapter, roy2023coprompt, seputis2024mma} to address BNT problem by incorporating multi-scale feature mixing. Recently, methods involving the introduction of LLMs \cite{zhang2023cafo, wang2024hpt} or knowledge distillation based on unlabeled images \cite{li2024promptkd, mistretta2024kdpl} are also explored to enhance generalization through external data.

Although the aforementioned methods alleviate the BNT problem to some extent, we believe that there is still a common limitation in existing research: the tuning approaches are all applied to the same set of prompt vectors, potentially facing interference between the optimization directions of base and new classes. In contrast, our Dual-Prompt Collaboration strategy decouples the optimization processes on base and new at the prompt level, providing a more fundamental approach to effectively overcome the BNT problem.

\section{Proposed Method}

The framework of \texttt{DPC} is demonstrated in \cref{Figure 3}. As a plug-and-play enhancement method, for an obtained pre-trained prompt tuning backbone, we first continuously optimize the newly established parallel prompt (\S \ref{Sec3.2}) on base classes through Dynamic Hard Negative Optimizer (\S \ref{Sec3.3}). Subsequently, during the inference phase, we employ the Weighting-Decoupling module (\S \ref{Sec3.4}) to perform decoupled generalization tasks for base and new classes based on dual prompts. The details of \texttt{DPC} are introduced as follows.

\subsection{Preliminaries}
Identical to extant research on prompt tuning, \texttt{DPC} utilizes CLIP as the pre-trained VLM backbone model for image-text feature extraction and modality interaction. CLIP employs ``A photo of a [CLASS]'' as the prompt template for the text modality and imports ViT-based \cite{dosovitskiy2020vit} visual encoder $f(\cdot)$ and text encoder $g(\cdot)$ to transform image $V$ and text $T$ into patch embedding and word embedding, respectively. During zero-shot inference, for a set of candidate objects $\boldsymbol{C}=\left\{T_{i}\right\}_{i=1}^{n}$, the matching probability between image features $f(V)$ and text features $g(T)$ is given by:
\begin{equation}
    p(y \mid V)=\frac{\exp \left(sim(g(T_y), f(V)) / \tau\right)}{\sum_{i=1}^{n} \exp \left(sim(g(T_i), f(V))/ \tau\right)}
\end{equation}
where $sim(\cdot, \cdot)$ denotes cosine similarity and $\tau$ is introduced as a temperature coefficient.

For transferring the foundation CLIP model to particular downstream tasks, prompt learner freezes the encoders $f(\cdot)$ and $g(\cdot)$, and appends a set of learnable vectors of length $M$ to the textual or visual inputs, typically organized as:
\begin{equation}
    \boldsymbol{P}=[\mathrm{p}]_{1}[\mathrm{p}]_{2}\ldots[\mathrm{p}]_{M}[CLASS]
\end{equation}

In most existing studies, text prompts $\boldsymbol{P}_t$ replace original prompt template of CLIP as the input for text modality. Optional visual prompts $\boldsymbol{P}_v$ are commonly joined as prefix to visual modality, concatenated with image patch tokens as $(\boldsymbol{P}_v,V)$. During tuning process, cross-entropy loss is normally applied to continuously optimize prompt vectors:
\begin{equation}
    \mathcal{L}_{\mathrm{CE}}=-\sum_{i} {h}_{i} \log p_{\theta}\left(y \mid V\right)
\end{equation}
\begin{equation}
    {p_\theta}(y \mid V)=\frac{\exp \left(sim(g({\boldsymbol{P}_t}_y), f(\boldsymbol{P}_v,V)) / \tau\right)}{\sum_{i=1}^{n} \exp \left(sim(g(\boldsymbol{P}_{t_i}), f(\boldsymbol{P}_v,V))/ \tau\right)}
\end{equation}
where ${h}_{i}$ is the one-hot label of the candidate object set $\boldsymbol{C}$.

\subsection{Dual Prompt Initialization} \label{Sec3.2}
In the initialization process of \texttt{DPC}, as a two-step tuning, we first execute moderate fine-tuning on the original prompt vector in prompt tuning backbone model, entirely adhering to the baseline settings to obtain the tuned prompt $\boldsymbol{P}$. Next, we freeze the tuned prompt and establish a set of learnable parallel prompt vectors $\boldsymbol{P}'$ based on it, with the form, size, and parameters cloned from the backbone model.
\begin{equation}
    \boldsymbol{P}' \coloneqq \boldsymbol{P}
\end{equation}

The dual prompts are designed to separately store latent features specific to base and new classes, thus decoupling the tasks for base and new at the prompt level. The frozen tuned prompt $\boldsymbol{P}$ is applied to guarantee generalization during new-class inference, while the parallel prompt $\boldsymbol{P}'$ is utilized for deeper optimization specific to base classes during fine-tuning stage. Detailed pipelines of backbone models that \texttt{DPC} used are listed in \emph{Supplementary Material \ref{sec:A2}}.

It is noteworthy that if the epochs or time length of fine-tuning are strictly limited, the tuning epochs on the backbone model can be halved, with the latter half replaced by fine-tuning based on the \texttt{DPC} optimizer. Through ablation experiments in \cref{Sec4.3}, we demonstrate that this setup can still achieve equivalent overall performance improvements for \texttt{DPC} without increasing computational cost.

\subsection{Dynamic Hard Negative Optimizer} \label{Sec3.3}
Independent of the original optimization process of the backbone, this module continuously fine-tunes the parallel prompt $\boldsymbol{P}^{\prime}$ by constructing a more challenging optimization task on base, effectively enhancing base class performance. It consists of three sub-modules: \emph{Negative Sampler}, \emph{Feature Filtering} and \emph{Hard Negative Optimizing}.

\noindent \textbf{Negative Sampler.} 
Replacing the random sampling strategy used by the backbone model, the Negative Sampler encourages the model to construct mini-batches utilizing hard negative samples that are tough to classify accurately. By reinforcing the difficulty of sample matching, the parallel prompt can be facilitated to fit the base class with tuning.

As the distribution of data shifts from zero-shot to base classes, the Top-$K$ results inferred by the fine-tuned prompt learner generally exhibit more approximate semantics on base tasks. We verify this character in \emph{Supplementary Material \ref{sec:B3}}. Therefore, for the ground-truth image-text pairs $(I_g, T_g)$ in the original mini-batch, we directly reuse the prompt tuning backbone, applying the frozen tuned prompt $\boldsymbol{P}$ as a query to dynamically obtain the Top-$K$ inference results, and treat the $K-1$ samples other than the positive object $T_g$ as hard negative objects $T^-$.

As subsequent process, hard negative objects and the positive object are concatenated to serve as the labels of the updated mini-batch $\boldsymbol{C}'=\left\{{T_g,T_{j}^{-}}\right\}_{j=1}^{K-1}$. Internal filtering is performed to exclude any identical objects within the mini-batch, and images matching the corresponding negative objects $\left\{{V_{j}^{-}}\right\}_{j=1}^{K-1}$ are randomly sampled from the training set to accommodate the following contrastive learning task. This process finally yields dynamic image-text pairs with size  $L\leq b \cdot K$ , where $b$ denotes the batch size.

It is crucial that to avoid data leakage, the Top-$K$ candidates of the negative sampler only contain base classes, and the image sampling range for constructing sample pairs is also restricted to the prebuilt train split. Compared to other hard negative samplers \cite{song2023hard, xie2023negative}, \texttt{DPC} achieves fully autonomous sample filtering without introducing any additional network parameters or external knowledge.

\noindent \textbf{Feature Filtering.} 
To maintain the complete performance of backbones, for obtaining the text features $g(\boldsymbol{C}')\in \mathbb{R}^{L\times d}$ generated by the text encoder from the set of hard negative objects $\boldsymbol{C}'$, \texttt{DPC} first performs L2 normalization on the text modality $\boldsymbol{C}$ during fine-tuning, which is constructed by the parallel prompt $\boldsymbol{P}'$ from all candidates $n$ in base classes. The purpose is to keep the global feature distribution of prompt learner for base classes unchanged, preventing parameter shift when collaborating with the tuned prompt $\boldsymbol{P}$.
\begin{equation}
    \hat{g}(\boldsymbol{C})=\frac{g(\boldsymbol{C})}{\|g(\boldsymbol{C})\|_{2}}\in \mathbb{R}^{n\times d},\ \  \boldsymbol{C}=\left\{{T}_{i}\right\}_{i=1}^{n}
\end{equation}

Next, a selection matrix $Q\in \mathbb{R}^{L\times n}$ is introduced for extracting text features associated with hard negatives.
\begin{equation}
    g(\boldsymbol{C}')= Q \cdot \hat{g}(\boldsymbol{C})
\end{equation}

$Q$ can be expressed as follows. ${\boldsymbol{e}_{i_{j}}}$ is the standard basis vector in $\boldsymbol{C}'$ where the label index position $i_j$ corresponding to the $j$-th hard negative object is 1 and the rest are 0.
\begin{equation}
    Q=\left(\mathbf{e}_{i_{1}},\  \mathbf{e}_{i_{2}},\  \ldots, \ \mathbf{e}_{i_{L}}\right), \ \ i \in \boldsymbol{C}^{\prime}
\end{equation}

With Feature Filtering, \texttt{DPC} reorganizes the image and text features input to the Hard Negative Optimizing process for subsequent optimization.

\begin{figure*}[t]
  \centering
  \includegraphics[width=\textwidth]{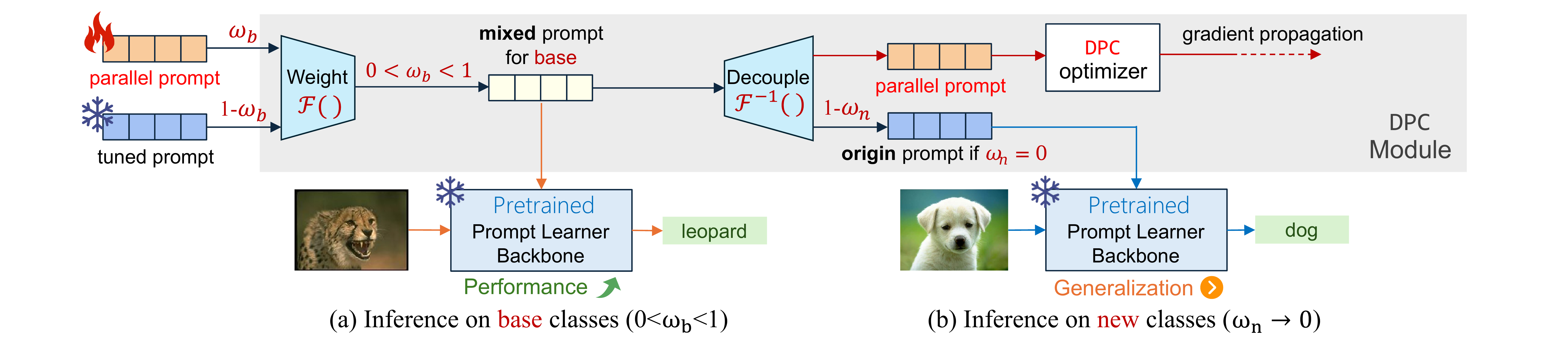}
  \caption{Weighting-Decoupling structure in \texttt{DPC}. This structure allows \texttt{DPC} to continuously optimize the parallel prompt $\boldsymbol{P}'$ during the tuing phase and to endow separate accumulated weights to dual prompts ($\boldsymbol{P}$ and $\boldsymbol{P}'$) during (a) inference stage on base classes and (b) inference stage on new classes.}
  \label{Figure 4}
\end{figure*}

\noindent \textbf{Hard Negative Optimizing.} 
To achieve more robust cross-modal alignment on base classes, we upgrade the cross-entropy loss of the traditional prompt learner to stronger image-text contrastive loss for hard negatives. For a mini-batch $(V',\boldsymbol{C}')$ composed of hard negative image-text pairs with length $L$, we employ the InfoNCE loss function \cite{wu2021infonce} to create a symmetric image-text contrastive learning task:
\begin{equation}
    \mathcal{L}_{\mathrm{CL}}=-\frac{1}{L} \sum_{i=1}^{L}\left(\log p_{\theta}\left(y \mid \boldsymbol{C}_{i}^{\prime}\right ) + \log p_{\theta}\left(y \mid V_{i}^{\prime}\right)\right)
\end{equation}

Among them, $p_{\theta}\left(y \mid \boldsymbol{C}_{i}^{\prime}\right)$ represents the matching score of the text feature for the target image $V_{i}^{\prime}$, and vice versa.

\begin{equation}
    p_{\theta}(y \mid \mathbf{C}_{i}^{\prime}) = \frac{\exp \left(\operatorname{sim}\left(f\left(V_{i}^{\prime}\right), g\left(\boldsymbol{C}_{i}^{\prime}\right)\right) / \tau\right)}{\sum_{j=1}^{L} \exp \left(\operatorname{sim}\left(f\left(V_{i}^{\prime}\right), g\left(\boldsymbol{C}_{j}^{\prime}\right)\right) / \tau\right)}
\end{equation}
\begin{equation}
    p_{\theta}(y \mid V_{i}^{\prime}) = \frac{\exp \left(\operatorname{sim}\left(g\left(\boldsymbol{C}_{i}^{\prime}\right), f\left(V_{i}^{\prime}\right)\right) / \tau\right)}{\sum_{j=1}^{L} \exp \left(\operatorname{sim}\left(g\left(\boldsymbol{C}_{i}^{\prime}\right), f\left(V_{j}^{\prime}\right)\right) / \tau\right)}
\end{equation}

If other losses are appended to the backbone model, $\mathcal{L}_{\mathrm{CL}}$ is computed in parallel as a plug-and-play optimizer. We believe that the above setups can benefit both visual prompts and textual prompts during optimization.

Overall, the Dynamic Hard Negative Optimizer enables parallel prompt $\boldsymbol{P}^{\prime}$ to better fit the base classes. Relying on the decoupled design of dual prompts, \texttt{DPC} optimizer does not compromise the generalization ability for new classes.

\subsection{Weighting-Decoupling Module} \label{Sec3.4}

Weighting-Decoupling Module (WDM) integrates the tuning and inference processes, allowing the tuned prompt $\boldsymbol{P}$ and parallel prompt $\boldsymbol{P}^{\prime}$ to decouple and collaborate flexibly.

WDM uniformly acts on the input of dual prompts during both tuning and inference stage. As demonstrated in \cref{Figure 4} (a), during model initialization, the Weighting sub-module $\mathcal{F}()$ is introduced, which combines the tuned prompt and parallel prompt into a mixed prompt $\widetilde{\boldsymbol{P}}_{b}$ by controlling the base-class-specific weighting coefficient $\omega_{b}$ constructed for base class inference.
\begin{equation}
    \widetilde{\boldsymbol{P}}_{b} = \mathcal{F}(\boldsymbol{P}^{\prime}) = \omega_{b} \boldsymbol{P}^{\prime}+\left(1-\omega_{b}\right) \boldsymbol{P}
\end{equation}

Subsequently, the mixed prompt is passed into the Decoupling sub-module, which is the inverse transformation of the Weighting module. As illustrated in \cref{Figure 4} (b), during this process, the mixed prompt is decomposed back into parallel prompt $\boldsymbol{P}^{\prime}$ and original tuned prompt $\boldsymbol{P}$ by $\mathcal{F}^{-1}()$. The former is imported to the \texttt{DPC} optimizer in tuning process to realize integrated gradient propagation. In contrast, during new class inference, both are reassigned by a new-class-specific weighting coefficient $\omega_{n}$ to obtain $\widetilde{\boldsymbol{P}}_{n}$:
\begin{equation}
    \widetilde{\boldsymbol{P}}_{n}=\omega_{n} \mathcal{F}^{-1}(\widetilde{\boldsymbol{P}}_{b})+\left(1-\omega_{n}\right) \boldsymbol{P}
\end{equation}

Above-mentioned design guarantees the model integrity while allowing independent weighting coefficients applied for base and new class inference, flexibly balancing the base class performance optimized by the parallel prompt $\boldsymbol{P}^{\prime}$ and the latent features for new class generalization of the tuned prompt $\boldsymbol{P}$. We discuss the range of $\omega_{b}$ and $\omega_{n}$ by ablation study in Section \ref{Sec4.3}.

\section{Experiments}

\subsection{Experimental Setup}  \label{Sec4.1}

\noindent \textbf{Datasets.} Following the benchmark setting of CoOp \cite{zhou2022coop}, for tasks of base-to-new generalization and cross-dataset transfer, we use 11 recognition-related datasets with diverse data distributions, including ImageNet \cite{deng2009imagenet}, Caltech101 \cite{fei2004caltech}, OxfordPets \cite{parkhi2012pets}, StanfordCars \cite{krause2013cars}, Flowers102 \cite{nilsback2008flowers}, Food101 \cite{bossard2014food}, FGVCAircraft \cite{maji2013aircraft}, SUN397 \cite{xiao2010sun}, DTD \cite{cimpoi2014dtd}, EuroSAT \cite{helber2019eurosat} and UCF101 \cite{soomro2012ucf101}. For cross-domain tasks, we select ImageNet-V2 \cite{recht2019imagenetv2}, ImageNet-Sketch \cite{wang2019imagenet-s}, ImageNet-A \cite{hendrycks2021imagenet-a} and ImageNet-R \cite{hendrycks2021imagenet-r}, which exhibit domain shifts compared to ImageNet.

\noindent \textbf{Baselines.} We select 4 influential prompt learners as baselines and backbone models for our plug-and-play module. These contain CoOp \cite{zhou2022coop} using \emph{textual} prompts, MaPLe \cite{khattak2023maple} employing \emph{integrated visual and textual} prompts, and PromptSRC \cite{khattak2023promptsrc} and PromptKD \cite{li2024promptkd}, which utilize \emph{separate visual and textual} prompts. Additionally, we compare the leading plug-and-play module for prompt learners, DePT \cite{zhang2024dept}, to validate the superiority of the prompt-level decoupling strategy of our \texttt{DPC}.


\begin{table*}[t!]
    \centering
    \renewcommand{\arraystretch}{0.8}
\begin{tabular}{c|ccc|ccc|ccc|ccc} 
\toprule
\multirow{2.4}{*}{Method} & \multicolumn{3}{c|}{\textbf{Avg. over 11 datasets}} & \multicolumn{3}{c|}{ImageNet}   & \multicolumn{3}{c|}{Caltech101} & \multicolumn{3}{c}{OxfordPets}    \\ 
\cmidrule(lr){2-4}\cmidrule(lr){5-7}\cmidrule(lr){8-10}\cmidrule(lr){11-13}
                        & Base  & New   & H                            & Base  & New   & H              & Base  & New   & H              & Base  & New   & H                 \\ 
\midrule
CoOp                    & 81.98 & 68.84 & 74.84                        & 76.41 & 68.85 & 72.43          & 97.55 & 94.65 & 96.08          & 95.06 & 97.60 & 96.31             \\
\cellcolor{cyan!15}\textbf{+DPC} & \cellcolor{cyan!15}\textbf{85.15} & \cellcolor{cyan!15}\textbf{68.84} & \cellcolor{cyan!15}\textbf{76.13}                        & \textbf{77.72} & \textbf{68.85} & \textbf{73.02}          & \textbf{98.58} & \textbf{94.65} & \textbf{96.58}          & \textbf{95.80} & \textbf{97.60} & \textbf{96.69}             \\ 
\midrule
MaPLe                   & 83.52 & 73.31 & 78.08                        & 76.91 & 67.96 & 72.16          & 97.98 & 94.50 & 96.21          & 95.23 & 97.67 & 96.44             \\
\cellcolor{cyan!15}\textbf{+DPC} & \cellcolor{cyan!15}\textbf{85.93} & \cellcolor{cyan!15}\textbf{73.31} & \cellcolor{cyan!15}\textbf{79.12}                        & \textbf{77.94} & \textbf{67.96} & \textbf{72.61}          & \textbf{98.64} & \textbf{94.50} & \textbf{96.53}          & \textbf{95.82} & \textbf{97.67} & \textbf{96.73}             \\ 
\midrule
PromptSRC               & 83.45 & 74.78 & 78.87                        & 77.28 & 70.72 & 73.85          & 97.93 & 94.21 & 96.03          & 95.41 & 97.30 & 96.34             \\
\cellcolor{cyan!15}\textbf{+DPC} & \cellcolor{cyan!15}\textbf{86.10} & \cellcolor{cyan!15}\textbf{74.78} & \cellcolor{cyan!15}\textbf{80.04}                        & \textbf{78.48} & \textbf{70.72} & \textbf{74.40}          & \textbf{98.90} & \textbf{94.21} & \textbf{96.50}          & \textbf{96.13} & \textbf{97.30} & \textbf{96.71}             \\ 
\midrule
PromptKD                & 86.86 & 80.55 & 83.59                        & \textbf{80.82} & 74.66 & \textbf{77.62}          & \textbf{98.90} & 96.29 & \textbf{97.58}          & \textbf{96.44} & 97.99 & \textbf{97.21}             \\
\cellcolor{cyan!15}\textbf{+DPC} & \cellcolor{cyan!15}\textbf{87.55} & \cellcolor{cyan!15}\textbf{80.55} & \cellcolor{cyan!15}\textbf{83.91}                        & 80.25 & \textbf{74.66} & 77.35          & 98.77 & \textbf{96.29} & 97.51          & 96.07 & \textbf{97.99} & 97.02             \\ 
\midrule\midrule
\multirow{2.4}{*}{Method} & \multicolumn{3}{c|}{StanfordCars}             & \multicolumn{3}{c|}{Flowers102} & \multicolumn{3}{c|}{Food101}    & \multicolumn{3}{c}{FGVCAircraft}  \\ 
\cmidrule(lr){2-4}\cmidrule(lr){5-7}\cmidrule(lr){8-10}\cmidrule(lr){11-13}
                        & Base  & New   & H                            & Base  & New   & H              & Base  & New   & H              & Base  & New   & H                 \\ 
\midrule
CoOp                    & 75.69 & 70.14 & 72.81                        & 96.96 & 68.37 & 80.19          & 90.49 & 91.47 & 90.98          & 37.33 & 24.24 & 29.39             \\
\textbf{+DPC}                    & \textbf{81.13} & \textbf{70.14} & \textbf{75.24}                        & \textbf{98.86} & \textbf{68.37} & \textbf{80.84}          & \textbf{91.15} & \textbf{91.47} & \textbf{91.31}          & \textbf{45.56} & \textbf{24.24} & \textbf{31.64}             \\ 
\midrule
MaPLe                   & 77.63 & 71.21 & 74.28                        & 97.03 & 72.67 & 83.10          & 89.85 & 90.47 & 90.16          & 40.82 & 34.01 & 37.11             \\
\textbf{+DPC}                    & \textbf{79.56} & \textbf{71.21} & \textbf{75.15}                        & \textbf{98.20} & \textbf{72.67} & \textbf{83.53}          & \textbf{91.35} & \textbf{90.47} & \textbf{90.90}          & \textbf{49.78} & \textbf{34.01} & \textbf{40.41}             \\ 
\midrule
PromptSRC               & 76.34 & 74.98 & 75.65                        & 97.06 & 73.19 & 83.45          & 90.83 & 91.58 & 91.20          & 39.20 & 35.33 & 37.16             \\
\textbf{+DPC}                    & \textbf{82.28} & \textbf{74.98} & \textbf{78.46}                        & \textbf{97.44} & \textbf{73.19} & \textbf{83.59}          & \textbf{91.40} & \textbf{91.58} & \textbf{91.49}          & \textbf{46.74} & \textbf{35.33} & \textbf{40.24}             \\ 
\midrule
PromptKD                & 82.41 & 82.80 & 82.60                        & \textbf{99.24} & 82.91 & \textbf{90.34}          & \textbf{92.59} & 93.73 & \textbf{93.15}          & 48.80 & 41.75 & 45.00             \\
\textbf{+DPC}                    & \textbf{84.17} & \textbf{82.80} & \textbf{83.48}                        & 98.96 & \textbf{82.91} & 90.23          & 92.41 & \textbf{93.73} & 93.07          & \textbf{52.94} & \textbf{41.75} & \textbf{46.68}             \\ 
\midrule\midrule
\multirow{2.4}{*}{Method} & \multicolumn{3}{c|}{SUN397}                   & \multicolumn{3}{c|}{DTD}        & \multicolumn{3}{c|}{EuroSAT}    & \multicolumn{3}{c}{UCF101}        \\ 
\cmidrule(lr){2-4}\cmidrule(lr){5-7}\cmidrule(lr){8-10}\cmidrule(lr){11-13}
                        & Base  & New   & H                            & Base  & New   & H              & Base  & New   & H              & Base  & New   & H                 \\ 
\midrule
CoOp                    & 80.99 & 74.10 & 77.39                        & 80.09 & 49.88 & 61.47          & 87.60 & 51.62 & 64.96          & 83.66 & 66.31 & 73.98             \\
\textbf{+DPC}                    & \textbf{82.81} & \textbf{74.10} & \textbf{78.21}                        & \textbf{84.61} & \textbf{49.88} & \textbf{62.76}          & \textbf{93.40} & \textbf{51.62} & \textbf{66.49}          & \textbf{87.02} & \textbf{66.31} & \textbf{75.27}             \\ 
\midrule
MaPLe                   & 81.54 & 75.93 & 78.63                        & 82.18 & 55.63 & 66.35          & 94.96 & 72.19 & 82.02          & 84.55 & 74.15 & 79.01             \\
\textbf{+DPC}                    & \textbf{82.02} & \textbf{75.93} & \textbf{78.86}                        & \textbf{85.48} & \textbf{55.63} & \textbf{67.40}          & \textbf{98.33} & \textbf{72.19} & \textbf{83.26}          & \textbf{88.14} & \textbf{74.15} & \textbf{80.54}             \\ 
\midrule
PromptSRC               & 82.28 & 78.08 & 80.13                        & 83.45 & 54.31 & 65.80          & 92.84 & 74.73 & 82.80          & 85.28 & 78.13 & 81.55             \\
\textbf{+DPC}                    & \textbf{83.63} & \textbf{78.08} & \textbf{80.76}                        & \textbf{86.88} & \textbf{54.31} & \textbf{66.84}          & \textbf{96.25} & \textbf{74.73} & \textbf{84.13}          & \textbf{88.99} & \textbf{78.13} & \textbf{83.21}             \\ 
\midrule
PromptKD                & \textbf{83.53} & 81.07 & \textbf{82.28}                        & 85.42 & 71.01 & 77.55          & 97.20 & 82.35 & 89.16          & 90.12 & 81.50 & 85.59             \\
\textbf{+DPC}                    & 83.28 & \textbf{81.07} & 82.17                        & \textbf{87.73} & \textbf{71.01} & \textbf{78.49}          & \textbf{98.29} & \textbf{82.35} & \textbf{89.61}          & \textbf{90.18} & \textbf{81.50} & \textbf{85.62}             \\
\bottomrule
\end{tabular}
    \caption{Base-to-new generalization performance of 4 backbone models w/ or w/o our \texttt{DPC} on 11 datasets. Benefiting from the decoupling structure at prompt level, \texttt{DPC} achieves general base class performance improvements while fully retaining new class generalization.}
    \label{tab1}
\end{table*}


\begin{table*}[t]
    \centering
    \renewcommand{\arraystretch}{0.8}
\setlength\tabcolsep{5.5pt}
\begin{tabular}{c|c|cc||c|c|cc} 
\toprule
\multirow{3.5}{*}{Method} & \textbf{Source}   & \multicolumn{2}{c||}{\textbf{Target}}                                                                                                       & \multirow{3.5}{*}{Method} & \textbf{Source}   & \multicolumn{2}{c}{\textbf{Target}}                                                                                                        \\ 
\cmidrule(lr){2-2}\cmidrule(lr){3-4}\cmidrule(lr){6-6}\cmidrule(lr){7-8}
                        & ImageNet & \begin{tabular}[c]{@{}c@{}}Avg. of \\cross-dataset\end{tabular} & \begin{tabular}[c]{@{}c@{}}Avg. of \\cross-domain\end{tabular} &                         & ImageNet & \begin{tabular}[c]{@{}c@{}}Avg. of \\cross-dataset\end{tabular} & \begin{tabular}[c]{@{}c@{}}Avg. of \\cross-domain\end{tabular}  \\ 
\midrule
CoOp                    & 71.25    & 64.98                                                           & 60.31                                                          & PromptSRC               & 70.65    & 65.64                                                           & 60.58
\\
\rowcolor{gray!20}
\textbf{+DPC}                    & \textbf{71.80}    & \textbf{64.98}                                                           & \textbf{60.31}                                                          & \textbf{+DPC}                    & \textbf{71.42}    & \textbf{65.64}                                                           & \textbf{60.58}                                                           \\
\midrule
\ \ \ MaPLe  \ \ \               & 70.11    & 64.79                                                           & 60.11                                                          & PromptKD                & 72.42    & 70.77                                                           & 71.47                                                           \\
\rowcolor{gray!20}
\textbf{+DPC}                    & \textbf{71.36}    & \textbf{64.79}                                                           & \textbf{60.11}                                                          & \textbf{+DPC}                    & \textbf{74.43}    & \textbf{70.77}                                                           & \textbf{71.47}                                                           \\
\bottomrule
\end{tabular}
    \caption{Average performance of cross-dataset and cross-domain generalization tasks of 4 backbone models w/ or w/o our \texttt{DPC}.}
    \label{tab2}
\end{table*}


\begin{figure}[ht]
  \centering
  \includegraphics[width=\linewidth]{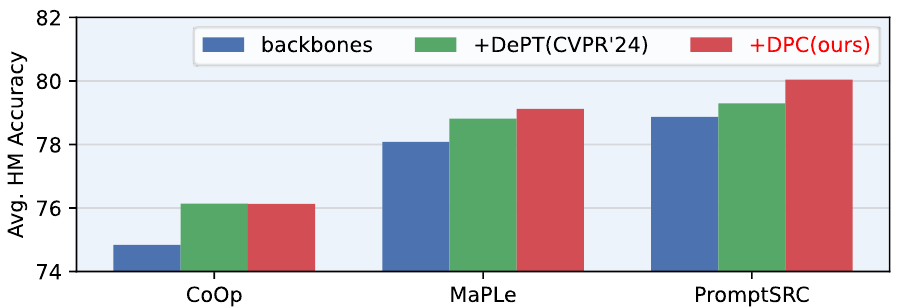}
  \caption{Average HM performance of base-to-new generalization tasks of 3 backbones with plug-and-play methods, DePT \cite{zhang2024dept} and our \texttt{DPC}.}
  \label{Figure 5}
\end{figure}

\noindent \textbf{Implementation Details.} We strictly follow the primary settings of the prompt tuning baselines, fine-tuning the backbone to obtain the tuned prompt, and subsequently fine-tuning the \texttt{DPC} optimizer utilizing the same hyperparameters. For a  fair comparison, we set the batch size of the backbone to 32, while for \texttt{DPC}, we select a batch size of 4 and set the Top-$K$ number of the Negative Sampler to $K=8$, ensuring that the size of the mini-batch remains consistent during fine-tuning. According to ablation study, the collaboration weights are set to $\omega_{b}=0.2$ (in MaPLe, $\omega_{b}=1.0$) and $\omega_{n}=1e$-6. Exceptionally, in PromptKD, due to its disparate settings (loading entire classes and fine-tuning based on all images in the dataset), we first maintain the original settings to obtain the tuned prompt. Next, the model is adjusted to sample few-shot image-text pair on base classes like other backbones, rendering the \texttt{DPC} optimizer learnable. Detailed implementation specifics of all models are enumerated in \emph{Supplementary Material \ref{sec:A2}}.

\begin{figure}[t]
  \centering
  \includegraphics[width=\linewidth]{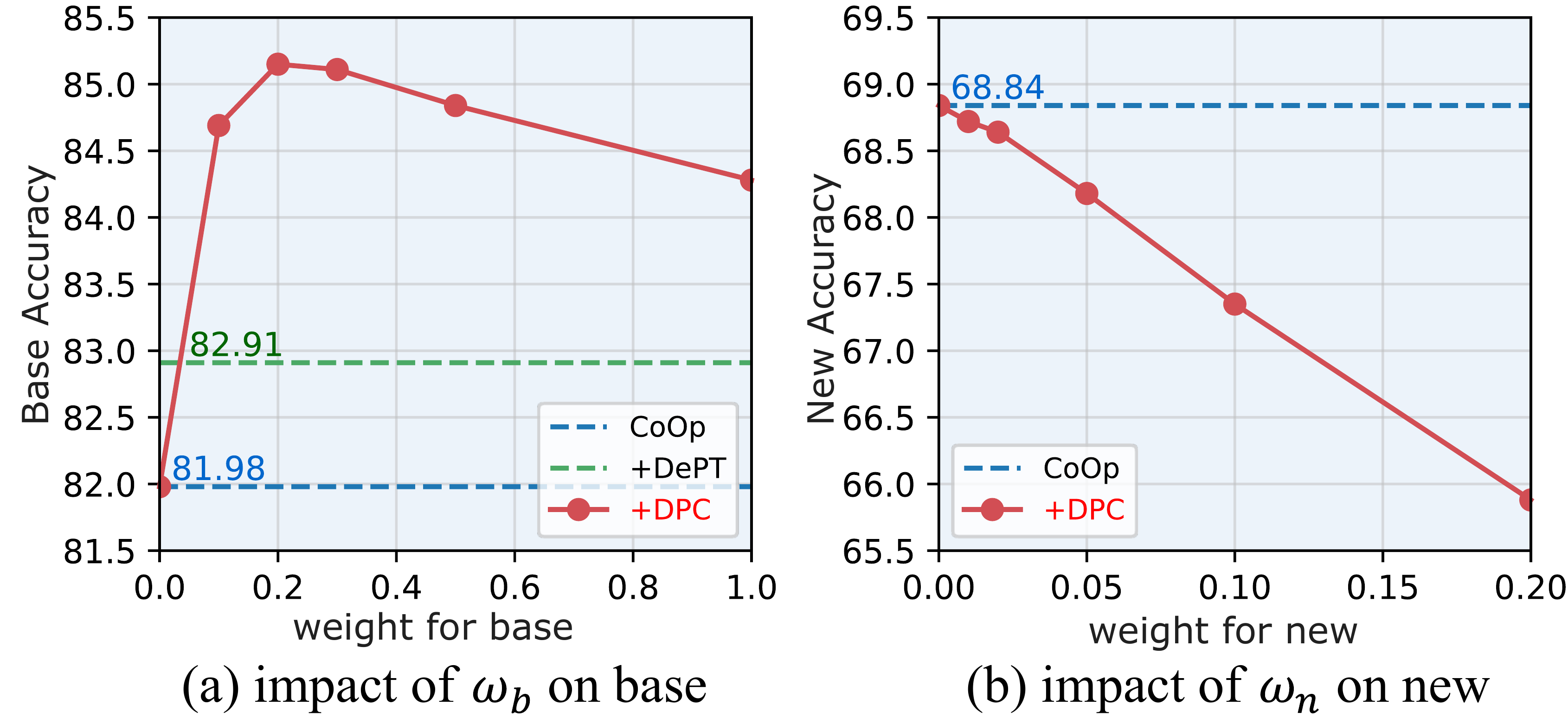}
  \caption{Impact of the collaboration weight for (a) base tasks and (b) new tasks in \texttt{DPC}. Detailed results are visible in Sup.Mat.B.2.}
  \label{Figure 6}
\end{figure}

\subsection{Experimental Results} \label{Sec4.2}

\noindent \textbf{Base-to-New Generalization.}
Adhering to the baselines, categories in each dataset are evenly divided into base classes and new classes. Fine-tuning of both the backbone and \texttt{DPC} is executed only on the base classes, followed by inference on both base and new classes. $H$ denotes the Harmonic Mean (HM) of the accuracy on base and new tasks. In conclusion, \texttt{DPC} achieves superior HM performance across all 4 backbones, while surpassing the current State-Of-The-Art PromptKD \cite{li2024promptkd} on multiple datasets. As exhibited in \cref{tab1}, the performance improvement mainly stems from the optimization on the base classes. Additionally, for tasks on new classes, the decoupled structure of \texttt{DPC} is validated to thoroughly maintain the generalization performance of the original backbone.

\noindent \textbf{Cross-Dataset and Cross-Domain Transfer.}
We train ImageNet on all classes as source and evaluate it on other datasets mentioned in \cref{Sec4.1} under zero-shot setting. Approximate to base-to-new tasks, \texttt{DPC} optimizes the performance on ImageNet, gaining enhancements across all backbones in \cref{tab2}. Meanwhile, for cross-dataset and cross-domain tasks involving unseen data distributions, the generalization level is consistently maintained through the coverage of raw tuned prompt $\boldsymbol{P}$.

\noindent \textbf{Compare with Another Plug-and-play Method.} 
To validate the effectiveness of \texttt{DPC} as a transferable module, we compare its HM performance with DePT \cite{zhang2024dept}, a plug-and-play prompt learner that decouples base and new tasks at the \textbf{feature} level, as displayed in \cref{Figure 5}. It is evident that the enhancement effect of \texttt{DPC} is superior or equal to DePT across the 3 baselines. We attribute this to the fact that decoupling at the \textbf{prompt} level is more thorough, furnishing a broader optimization space for the plug-and-play model. More detailed data is listed in \emph{Supplementary Material \ref{sec:B7}}.

\begin{table}
    \centering
    \renewcommand{\arraystretch}{0.8}
\setlength\tabcolsep{5pt}
\begin{tabular}{ccccc|ccc} 
\toprule
                         &                    TS & DHNO & WE & DE                   & Base  & New   & H                          \\ 
\midrule
                &                     &     &    &                      & 81.98 & 68.84 & 74.84                      \\
(1)                       &  \checkmark                   &     &    &                      & 82.69 & 68.39 & 74.86                      \\
(2)                       & \checkmark                    & \checkmark    &    &                      & 84.28 & 64.12 & 72.83                      \\
(3)                       & \checkmark                    & \checkmark    & \checkmark   &                      & 85.15 & 65.88 & 74.29                      \\
(4)                       & \checkmark                    &     & \checkmark   & \checkmark                     & 82.23 & 68.84 & 74.94                      \\
\rowcolor{gray!20}
(5)                       & \checkmark                    & \checkmark    & \checkmark   & \checkmark                     & \textbf{85.15} & \textbf{68.84} & \textbf{76.13}                      \\
\bottomrule
\end{tabular}
    \caption{Ablation study of components in \texttt{DPC} with CoOp baseline on base-to-new tasks. TS: Two-Step tuning. DHNO: Dynamic Hard Negative Optimizer. WE \& DE: Weighting-Decoupling.}
    \label{tab3}
\end{table}

\subsection{Ablation Study} \label{Sec4.3}

In this section, we discuss the impact of each \texttt{DPC} sub-modules, collaboration weights ($\omega_{b},\omega_{n}$), Top-$K$ sampling amount and fine-tuning epoch on model performance. The evaluation is based on base-to-new tasks of CoOp. More detailed experiments are listed in \emph{Supplementary Material}.

\noindent \textbf{Validity of Proposed Components.}
\cref{tab3} illustrates the performance variation by introducing \texttt{DPC} sub-modules into backbones. Comparison between \emph{(1)} tuning the original backbone continuously with another 20 epochs, \emph{(2)} introducing Dynamic Hard Negative Optimizer (DHNO), and \emph{(3)} performing dual-prompt weight accumulation (WE) on base classes, validates the effectiveness of DHNO and WE in enhancing base performance, respectively. However, the BNT problem can be observed, manifesting as a continuous decline of new-class performance. This suggests that methods without prompt-level decoupling tend to overfit to base classes. In contrast, \emph{(4)} model with complete Weighting-Decoupling successfully maintains generalization performance, highlighting its necessity. Nonetheless, the absence of DHNO results in limited promotion of base tasks. By comparison, \emph{(5)} with full configuration performs the best, confirming that both optimization directions for base and new classes are indispensable and proving the superiority of decoupling at the prompt level. Further ablation studies on DHNO sub-modules are in \emph{Supplementary Material \ref{sec:B5}}.

\noindent \textbf{Influence of Collaboration Weights.}
The impact of collaboration weights ($\omega_{b},\omega_{n}$) on base and new tasks is reflected in \cref{Figure 6}. Overall, \texttt{DPC} reaches optimal performance with $\omega_{b}$=0.2 and $\omega_{n}$=1$e$-6. By prompt decoupling, weights for base and new tasks are independently valued, avoiding the BNT problem in inference stage. Concrete data and analyses are detailed in \emph{Supplementary Material \ref{sec:B2}}.

\begin{table}
    \setlength\tabcolsep{4pt}
    \centering
    \renewcommand{\arraystretch}{0.8}
\begin{tabular}{cc|cc} 
\toprule
batch size & Top-$K$ & Avg. Accuracy      & Time   \\ 
\midrule
4  & 8    & 85.15 (\textcolor{V}{+3.17}) & 1X     \\
8  & 4    & 84.50 (\textcolor{V}{+2.52}) & 0.69X  \\
16 & 2    & 83.84 (\textcolor{V}{+1.85}) & 0.59X  \\
\bottomrule
\end{tabular}
    \caption{Ablation study of the amount of Top-$K$ in Negative Sampler. Size of mini-batch is fixed at 32 for fair comparison.}
    \label{tab4}
\end{table}
\begin{table}
    \centering
    \renewcommand{\arraystretch}{0.8}
\begin{tabular}{c|cc|ccc} 
\toprule
Model    & epoch  & total               & Base  & New   & H      \\ 
\midrule
backbone & 20  & \multirow{2}{*}{40} & 81.98 & 68.84 & 74.84  \\
+DPC    & +20 &                     & \textbf{85.15} & \textbf{68.84} & \textbf{76.13}  \\ 
\midrule
backbone & 10  & \multirow{2}{*}{20} & 81.68 & 70.75 & 75.82  \\
+DPC    & +10 &                     & \textbf{83.99} & \textbf{70.75} & \textbf{76.80}  \\ 
\midrule
backbone & 5   & \multirow{2}{*}{10} & 79.64 & 74.19 & 76.82  \\
+DPC    & +5  &                     & \textbf{82.89} & \textbf{74.19} & \textbf{78.29}  \\
\bottomrule
\end{tabular}
    \caption{Ablation study of the effect with less fine-tuning epoch.}
    \label{tab5}
\end{table}

\noindent \textbf{Impact of Top-$K$ Sampling Amount.}
Under the premise of a fixed mini-batch size $L$, we test diverse Top-$K$ sampling amounts of the Negative Sampler (\S \ref{Sec3.3}) and summarize the comparative results in \cref{tab4}. We observe that the base performance promotes with the growth of $K$, demonstrating that the Negative Sampler effectually collects and learns more similar hard negatives. From another perspective, affected by the data interaction bottleneck of the sampler, time of PEFT can be further reduced by decreasing $K$.

\noindent \textbf{Less Fine-tuning Epochs.}
As an approach to reduce the computational cost, we consider smaller total amounts of epochs. The strategy is discussed in Section \ref{Sec3.2}. As shown in \cref{tab5}, even when the epochs are cut back to half or a quarter, the base performance of \texttt{DPC} remains superior to the original backbone. Meanwhile, with the intensive generalization performance for new classes, HM performance growth relative to backbones can still be acquired.

\noindent \textbf{Computational Cost.} As discussed in \emph{Supplementary Material \ref{sec:B6}}, the additional computational cost of \texttt{DPC} is tiny. Compared to baselines, \texttt{DPC} does not introduce a significant increase in parameters, memory cost, or inference time.

\subsection{Interpretability and Analysis} \label{Sec4.4}
To provide a empirical analysis to the mechanism of the \texttt{DPC} weight accumulation structure, in this section, we demonstrate and analyze the feature channel invariance of the prompt vectors during fine-tuning that we discover.

\begin{figure}[t]
  \centering
  \includegraphics[width=\linewidth]{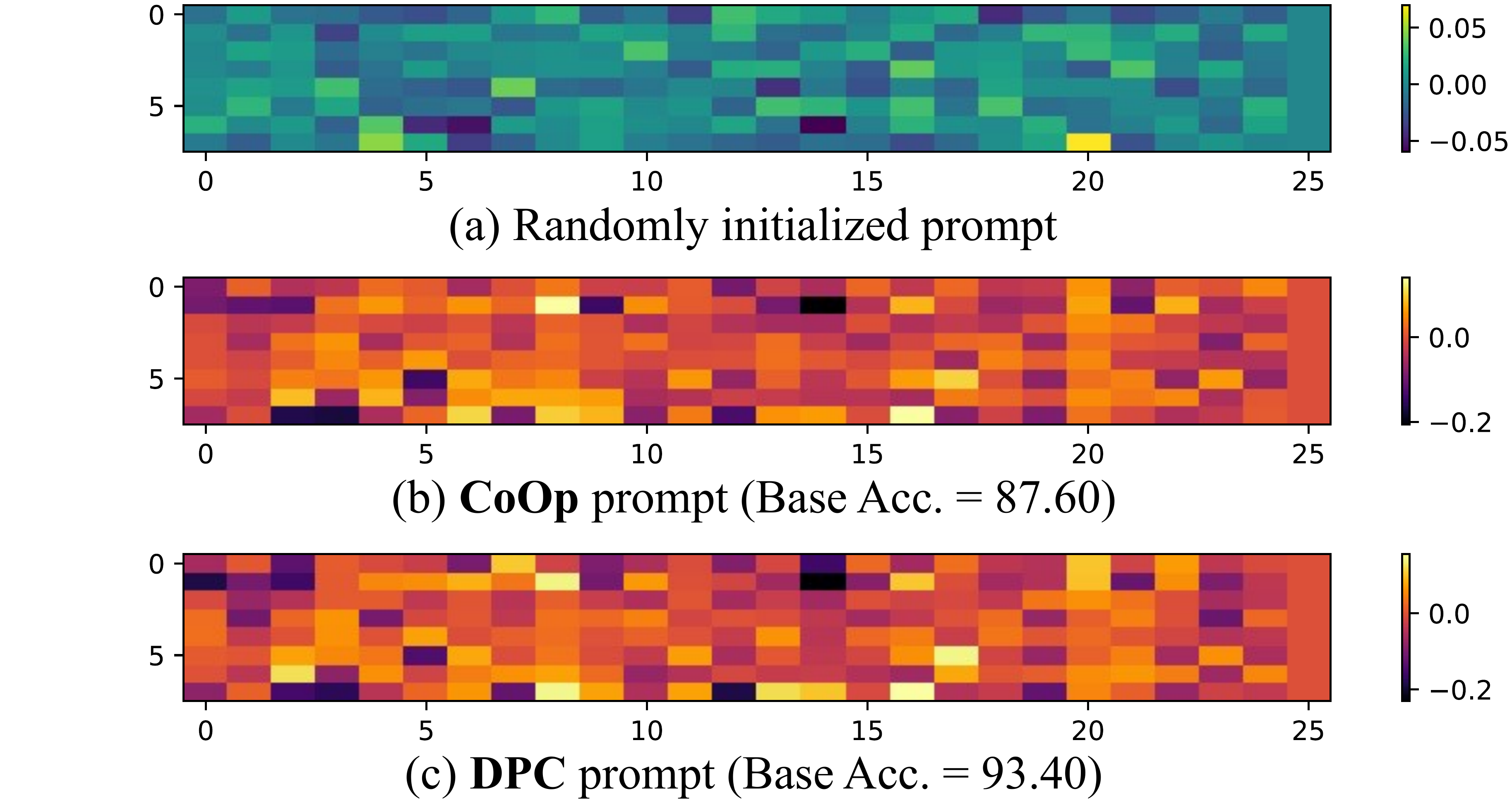}
  \caption{Visualization of feature maps of (a) randomly initialized prompt before tuning, (b) tuned prompt on CoOp backbone, and (c) optimized parallel prompt of \texttt{DPC}. Prompts are fine-tuned on base classes of EuroSAT \cite{helber2019eurosat} and are down-sampled for readability.}
  \label{Figure 7}
\end{figure}

As a visualization, we map the randomly initialized prompt vector, the tuned prompt $\boldsymbol{P}$ optimized by the backbone model, and the parallel prompt $\boldsymbol{P}^{\prime}$ obtained after \texttt{DPC} optimization onto the feature maps in \cref{Figure 7}. We find that the feature distribution of parallel prompt fine-tuned by \texttt{DPC} is highly similar to the original tuned prompt.

We reveal this phenomenon through the following analysis: During \texttt{DPC} optimization, the parallel prompt $\boldsymbol{P}^{\prime}$ is initialized based on the tuned prompt $\boldsymbol{P}$, and both are fine-tuned on the tasks targeting identical base classes, following the design of the decoupled structure (\S \ref{Sec3.4}). Benefiting from the Feature Filtering module (\S \ref{Sec3.3}) that maintains the original feature distribution of the base, the latent feature channels basically remain unchanged during the \texttt{DPC} optimization on parallel prompt. This characteristic allows \texttt{DPC} to linearly control the shift of the mixed prompt $\widetilde{\boldsymbol{P}}_{b}$ towards the base classes by dynamically adjusting weights $\omega_{b}$, thereby maximizing HM performance.

\section{Conclusion} \label{Sec5}
We propose \texttt{DPC}, the first approach that decoupling at \textbf{prompt} level to address the Base-New Trade-off problem in prompt tuning. During fine-tuning, the tuned prompt obtained from backbone is frozen to maintain generalization to new tasks, while also being applied as a query for Negative Sampler to spontaneously construct hard negatives for optimization. The activated parallel prompt significantly enhances base performance through the Dynamic Hard Negative Optimizer. During inference, by introducing decoupled weights for base and new, features from dual prompts are flexibly coordinated to maximize overall performance.

In future work, we will explore further improvements to \texttt{DPC}, such as adaptive parameterization of the weight coefficient $\omega$ and adaptation to a broader range of downstream tasks (e.g., object detection and semantic segmentation).

\section*{Acknowledgements}
This work is supported by grants from the Natural Science Foundation of Shanghai, China (No. 23ZR1422800), Shanghai Institute of Intelligent Science and Technology in Tongji University, China Scholarship Council (CSC) and UTS Top-Up Scholarship.

{
    \small
    \bibliographystyle{ieeenat_fullname}
    \bibliography{main}
}

\clearpage
\setcounter{page}{1}
\setcounter{section}{0}
\setcounter{subsection}{0}
\maketitlesupplementary
\appendix

\section{More Implementation Details} \label{sec:A}
Herein, we provide additional detailed setup of \texttt{DPC} to enhance the reproducibility of our model.

\subsection{Experimental Setup} \label{sec:A1}

\noindent \textbf{Datasets.}
As described in the main text, for most datasets, we restrict the size of mini-batch sampled by the Dynamic Hard Negative Optimizer to $L\leq32$ when executing \texttt{DPC} fine-tuning. The seed of \texttt{DPC} is consistent with backbone.

However, it is important to note that for the DTD \cite{cimpoi2014dtd} and OxfordPets \cite{parkhi2012pets} datasets, after the base and new splitting, there are only 24 and 19 sub-classes that involved in the base tasks, respectively, which are fewer than 32. Since the Dynamic Hard Negative Optimizer requires maintaining the size of the mini-batch smaller than the quantity of base classes (otherwise, the effectiveness of hard negative selection would be compromised), we reduce the parameters to $b=2$ and $K=8$ for these two datasets, ensuring $L\leq16$. Furthermore, since EuroSAT \cite{helber2019eurosat} possesses only 5 base classes, we set $b=2$ and $K=2$ during optimization on this dataset.

Apart from these, the data sampling strategy for the inference process and fine-tuning on the backbone remains consistent with the baselines.

\noindent \textbf{Hyperparemeters.} 
Following the setup of the backbones, we utilize the ViT-B/16-based CLIP as the foundation model for prompt learners. For a fair comparison, all backbones and \texttt{DPC} are fine-tuned for epochs $ep=20$ with learning rate $lr=0.002$ for base-to-new tasks to avoid gradient explosion, and a 16-shot setting is applied to all models except PromptKD backbone. For cross-dataset tasks, we follow the PromptSRC \cite{khattak2023promptsrc} settings, fine-tuning on all categories of ImageNet with $ep=5$ and a learning rate $lr=0.0035$, while reducing the depth of visual and text prompts to 3 (except for CoOp).

Detailed information of the text and visual prompt settings is enumerated in \cref{tab6}. For the initialization process, text prompt in CoOp is randomly initialized adhering a zero-mean Gaussian distribution, while the other 3 backbones apply the encoded ``A photo of a'' tokens as the initialization template. Additionally, PromptSRC and PromptKD follow the Independent Vision-Language Prompt (IVLP) \cite{rasheed2023ivlp} setting, where prompts related to the two modalities are independently initialized. We use 1 Tesla A40 GPU to perform 3 runs on each dataset.

\begin{figure}[th]
  \centering
  \includegraphics[width=0.97\linewidth]{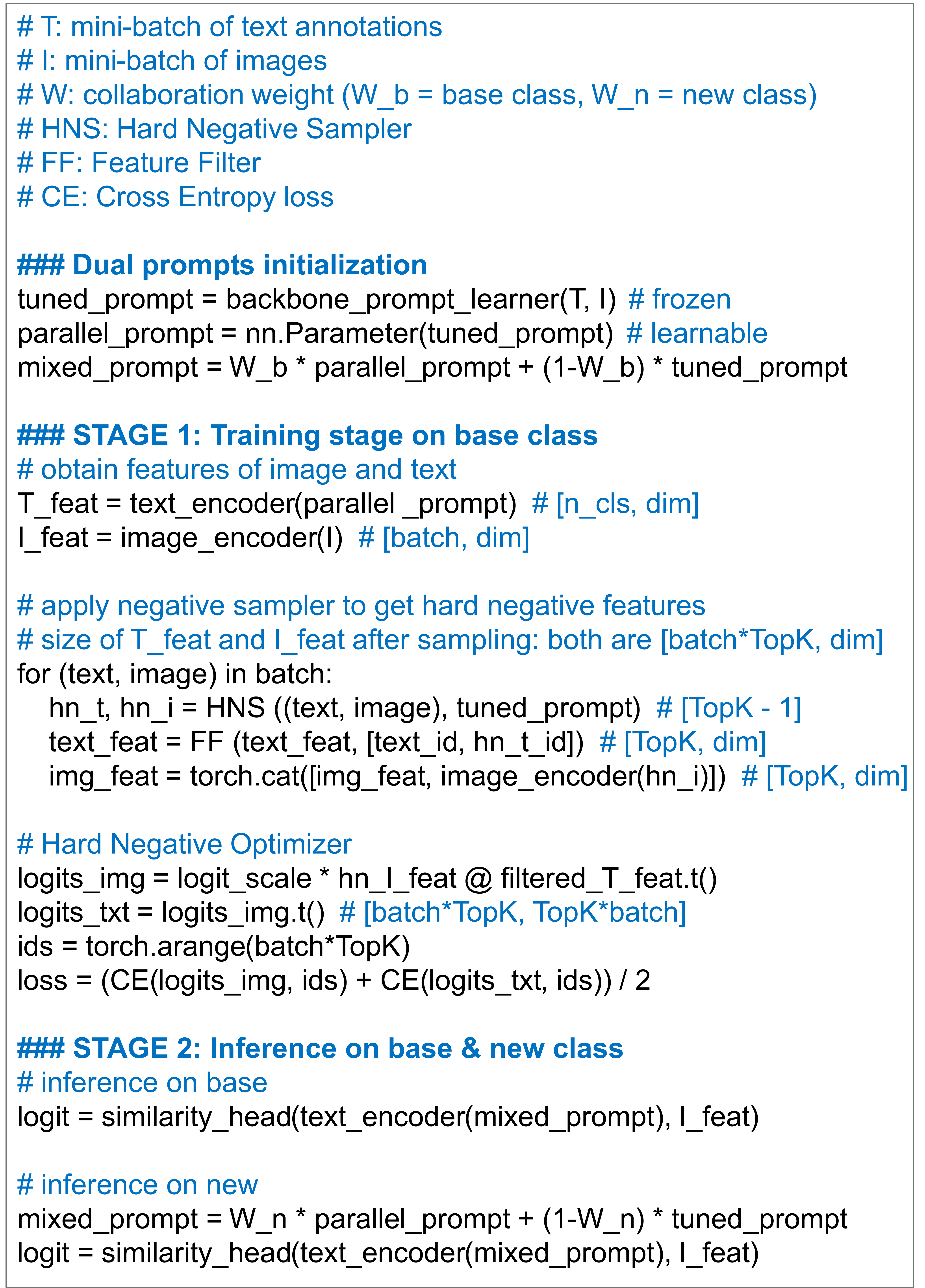}
  \caption{Pseudo-code of \texttt{DPC} in PyTorch. The size of dynamic hard negative mini-batch is considered as $L=b \cdot K$ for easier understanding.}
  \label{Figure 8}
\end{figure}
\begin{table}[th]
\centering
\setlength\tabcolsep{3.5pt}
\begin{tabular}{ccccc} 
\toprule
Params              & CoOp & MaPLe & ProSRC & ProKD  \\ 
\midrule
Text prompt depth   & 1    & 9     & 9         & 9         \\
Visual prompt depth & -    & 9     & 9         & 9         \\
Context length      & (4,0)    & (2,4) & (4,4)     & (4,4)     \\
Prompt layer        & 1    & 12    & 12        & 12        \\
Optimizer           & SGD  & SGD   & SGD       & SGD       \\
\bottomrule
\end{tabular}
    \caption{Training settings of backbones for base-to-new tasks.}
    \label{tab6}
\end{table}

\noindent \textbf{Algorithm.} In \cref{Figure 8}, we demonstrate the \texttt{DPC} procedure as pseudo-code. For clarity, although the mini-batch sampled by \texttt{DPC} through the Dynamic Hard Negative Optimizer has a variable size $L$, we annotate the tensor dimensions in the comments with the assumption $L = b\cdot K$, meaning that all the hard negative objects sampled in this mini-batch are non-repetitive. This hypothesis does not affect the actual process of the model.

\subsection{DPC Optimization for Backbones} \label{sec:A2}
As a robust plug-and-play module, the \texttt{DPC} Optimizer performs targeted modifications to various backbones based on separate forms of prompts and model architectures to achieve complete model adaptation. In this section, we provide a brief introduction to the frameworks of the 4 selected backbones and declare the specific strategies for introducing and fine-tuning the \texttt{DPC} module.

\noindent \textbf{CoOp \cite{zhou2022coop}.}
CoOp briefly introduces a randomly initialized text prompt to replace the original fixed template ``A photo of a [CLASS]''. Obeying the introduction of the \texttt{DPC} framework in the main text, we first fine-tune the original CoOp backbone to obtain the tuned text prompt $\boldsymbol{P}$. Subsequently, for the \texttt{DPC} optimizer, we append the parallel prompt $\boldsymbol{P}^{\prime}$ into the text modality for dual-prompt collaboration, while replacing the cross-entropy loss of CoOp with the contrastive learning loss in hard-negatives $\mathcal{L}_{\mathrm{CL}}$ of \texttt{DPC} for subsequent incremental fine-tuning.

\noindent \textbf{MaPLe \cite{khattak2023maple}.}
MaPLe integrates visual and text prompts by establishing a set of activated feature mapping layers, which derive corresponding visual prompts from learnable text prompts. Within the \texttt{DPC} framework, after fine-tuning the original backbone, we obtain sets of visual and text prompts ($\boldsymbol{P}_{vi}, \boldsymbol{P}_{ti}$) as initial values for the parallel prompts (${\boldsymbol{P}_{vi}}^{\prime}, {\boldsymbol{P}_{ti}}^{\prime}$) and load the weight parameters of the feature mapping layers to initialize the \texttt{DPC} optimizer. Since only text prompts $\boldsymbol{P}_{ti}$ are learnable in MaPLe, similar to CoOp, we upgrade the cross-entropy loss of MaPLe to \texttt{DPC} contrastive learning loss $\mathcal{L}_{\text{CL}}$ in subsequent stages, while continuously optimizing the text-based parallel prompts ${\boldsymbol{P}_{ti}}^{\prime}$ while keeping the mapping layers for visual prompts activated within \texttt{DPC}. In the Weighting-Decoupling weight accumulation module during the inference process, we apply the same base-class weights $\omega_{b}$ or new-class weights $\omega_{n}$ for prompts of both modalities.

\noindent \textbf{PromptSRC \cite{khattak2023promptsrc}.}
PromptSRC employs independent visual and text prompts for fine-tuning, following the IVLP setting. It introduces more robust loss functions as constraints to mitigate the negative impact of the BNT problem. Specifically, in addition to the cross-entropy loss $\mathcal{L}_{\text{CE}}$ adopted by typical prompt learners, PromptSRC also appends consistency constraints between the prompts and their corresponding modality features,  $\mathcal{L}_{\text{SCL-image}}$ and $\mathcal{L}_{\text{SCL-text}}$, as well as a further constraint between the logits after modality interaction, $\mathcal{L}_{\text{SCL-logits}}$, to balance the base-new performance.

\begin{figure}[t]
  \centering
  \includegraphics[width=\linewidth]{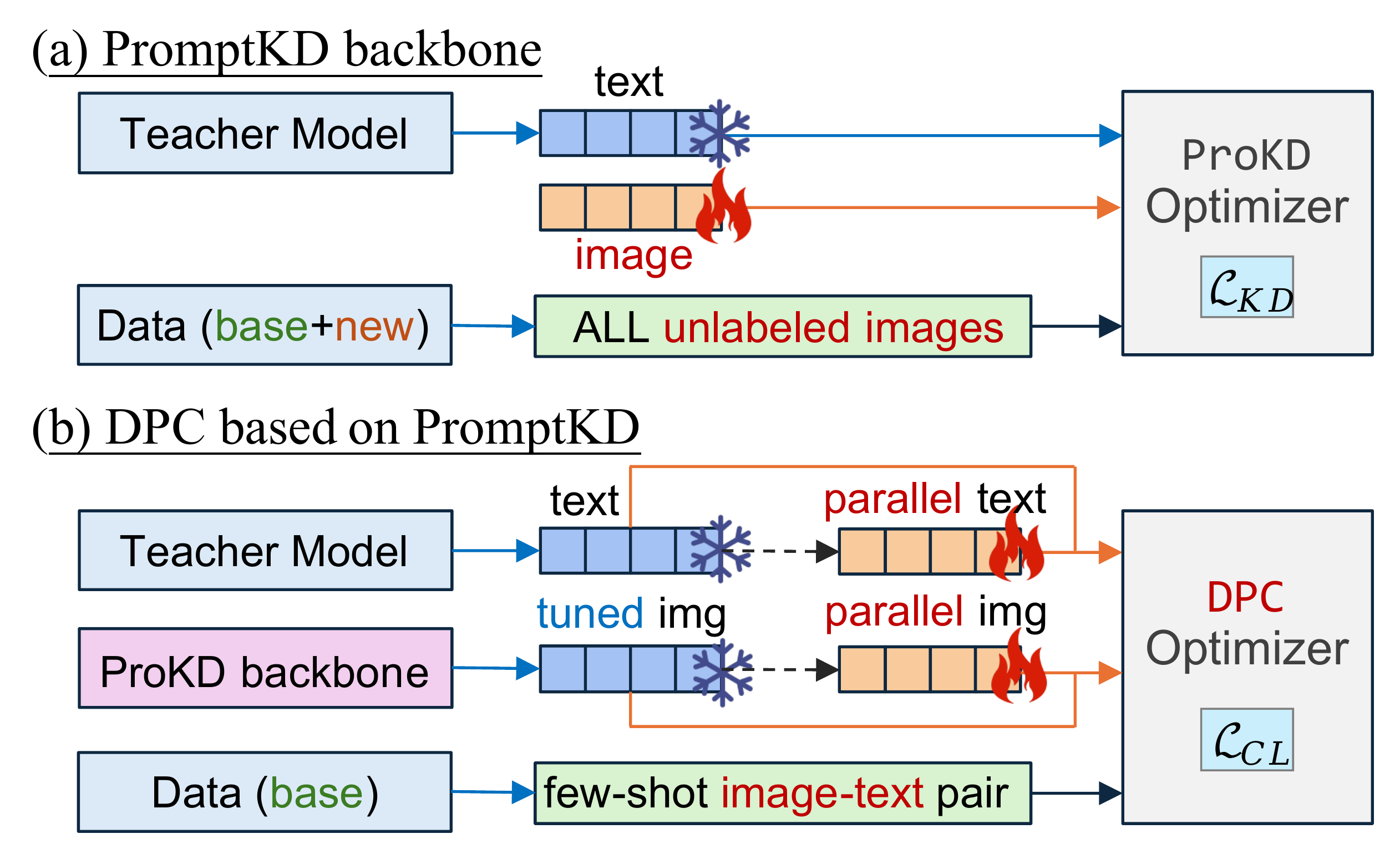}
  \caption{Initialization of text \& image prompts and optimizers in PromptKD backbone and \texttt{DPC}.}
  \label{Figure 9}
\end{figure}
\begin{table}[t]
\centering
\begin{tabular}{cccc} 
\toprule
\multirow{2.4}{*}{\textbf{Model}} & \multicolumn{3}{c}{\textbf{Avg. accuracy}}  \\ 
\cmidrule(lr){2-4}
                       & Base  & New   & H                 \\ 
\midrule
CLIP \cite{radford2021clip}                  & 69.34 & 74.22 & 71.70             \\
CoCoOp \cite{zhou2022cocoop}                & 80.47 & 71.69 & 75.83             \\
KgCoOp \cite{yao2023kgcoop}                & 80.73 & 73.60 & 77.00             \\
TCP \cite{yao2024tcp}                   & 84.13 & \textbf{75.36} & 79.51             \\
\rowcolor{gray!20}
\textbf{DPC-SRC}                & \textbf{86.10} & 74.78 & \textbf{80.04}             \\
\midrule
KDPL \cite{mistretta2024kdpl}                  & 77.11 & 71.61 & 74.26             \\
CoPrompt \cite{roy2023coprompt}                 & 84.00 & 77.23 & 80.48             \\
\rowcolor{gray!20}
\textbf{DPC-PK}                & \textbf{87.55} & \textbf{80.55} & \textbf{83.91}             \\
\bottomrule
\end{tabular}
    \caption{Comparison with additional prompt tuning baselines fine-tuned based on internal constraints or external knowledge for base-to-new tasks. \texttt{DPC-SRC} denotes a combination of DPC and PromptSRC, while \texttt{DPC-PK} is a binding of DPC and PromptKD.}
    \label{tab7}
\end{table}

Therefore, in the \texttt{DPC} framework, after obtaining the visual and text tuned prompts ($\boldsymbol{P}_{vi}, \boldsymbol{P}_{ti}$) optimized by the backbone, we construct parallel prompts (${\boldsymbol{P}_{vi}}^{\prime}, {\boldsymbol{P}_{ti}}^{\prime}$) based on both modalities, keeping them activated to sustain learnability. During \texttt{DPC} optimization, we replace the original $\mathcal{L}_{\text{CE}}$ with $\mathcal{L}_{\text{CL}}$ that corresponding to \texttt{DPC}, while the other 3 loss functions are directly inherited by the \texttt{DPC} optimizer, collectively contributing to the continuous fine-tuning of the parallel prompts. Similarly, during inference stage, the visual and text parallel prompts (${\boldsymbol{P}_{vi}}^{\prime}, {\boldsymbol{P}_{ti}}^{\prime}$) are still weight-accumulated with the corresponding tuned prompts in relevant modalities of PromptSRC backbone to achieve intra-modality dual-prompt collaboration.

\noindent \textbf{PromptKD \cite{li2024promptkd}.}
As a knowledge distillation-driven model, PromptKD introduces PromptSRC fine-tuned on larger ViT-L/14 as the teacher model. Unlike other backbones, PromptKD processes unlabeled images from the entire dataset and applies the teacher model to infer and optimize the student model across all base and new classes during fine-tuning. In this procedure, the text prompts $\boldsymbol{P}_{ti}$ extracted from the teacher model are frozen, while only the prompts in the visual branch $\boldsymbol{P}_{vi}$ and a projection layer $h(\cdot)$ for aligning the student model with the teacher model are updated.

To integrate the \texttt{DPC} optimizer into PromptKD, we devise a targeted framework, as shown in \cref{Figure 9}. Initially, we fine-tune the original PromptKD backbone to obtain the visual prompts $\boldsymbol{P}_{vi}$ and the parameters of the projection layer. Subsequently, we fully modify the Dataloader of PromptKD, altering it from loading unlabeled images across all categories to sampling few-shot image-text pairs from base classes, aligning it with other prompt tuning backbones.

Corresponding to the data input modification, fine-tuning strategy of PromptKD is also momentously updated to accommodate our \texttt{DPC} optimizer. Specifically, we construct parallel prompts (${\boldsymbol{P}_{vi}}^{\prime}, {\boldsymbol{P}_{ti}}^{\prime}$) based on the frozen text prompts from the teacher model $\boldsymbol{P}_{ti}$ and the visual prompts fine-tuned by PromptKD $\boldsymbol{P}_{vi}$, then set both of them activated. During \texttt{DPC} fine-tuning, all original loss functions of PromptKD are disabled, while only the \texttt{DPC} image-text contrastive loss $\mathcal{L}_{\text{CL}}$ is applied for further optimization. It is worth noticing that to maintain the original generalization performance of PromptKD, the contrastive loss is applied under the ViT-L/14 setting of teacher model, transmitting the text parallel prompts ${\boldsymbol{P}_{ti}}^{\prime}$ and the visual parallel prompts upscaled by the activated projection layer ${h(\boldsymbol{P}_{vi}}^{\prime})$ as inputs to the feature encoders. The weight accumulation procedure is consistent with \texttt{DPC} in PromptSRC.

In summary, being constructed as an independent task, \texttt{DPC} is introduced into PromptKD based on few-shot image-text data as a plug-and-play module. Aforementioned design successfully integrates \texttt{DPC} optimization while preserving the original performance of PromptKD.

\section{More Experimental Results} \label{sec:B}
Herein, we supplement the main text with more elaborated experiments. Performance comparisons with additional prompt tuning baselines (\S \ref{sec:B1}), the specific impact of collaboration weights ($\omega_{b},\omega_{n}$) on each dataset (\S \ref{sec:B2}), similarity measurements of samples from the Hard Negative Sampler (\S \ref{sec:B3}), the effects of the \texttt{DPC} optimizer on the visual or text branches (\S \ref{sec:B4}), more ablation studies on \texttt{DPC} components (\S \ref{sec:B5}), and assessments of computational cost (\S \ref{sec:B6}) are contained in this section.

\subsection{Compare with More Baselines} \label{sec:B1}
To further highlight the comprehensive performance advantages of \texttt{DPC}, more baselines are brought in for comparison on base-to-new generalization tasks. As illustrated in \cref{tab7}, we compare \texttt{DPC} based on PromptSRC (\texttt{DPC-SRC}) with the initial CLIP and other models optimized by internal constraints, containing CoCoOp \cite{zhou2022cocoop}, KgCoOp \cite{yao2023kgcoop}, and TCP \cite{yao2024tcp}. For knowledge distillation-based models, KDPL \cite{mistretta2024kdpl} and CoPrompt \cite{roy2023coprompt} are utilized for contrasting with the combination of \texttt{DPC} and PromptKD (\texttt{DPC-PK}). It is apparent that models reinforced by \texttt{DPC} surpass the current baselines, achieving the latest State-Of-The-Art performance.

\subsection{Detailed Ablation on Collaboration Weights} \label{sec:B2}

\begin{table}[th]
\setlength\tabcolsep{2.8pt}
\centering
\begin{tabular}{ccccccc}
\toprule
\multirow{2.4}{*}{\textbf{Dataset}} & \multicolumn{6}{c}{\textbf{weight for base class ($\omega_{b}$)}}      \\ 
\cmidrule(lr){2-7}
                         & 0     & 0.1   & 0.2   & 0.3   & 0.5   & 1      \\ 
\midrule
ImageNet                 & 76.41 & 77.72 & 77.72 & \textbf{77.95} & 77.92 & 77.58  \\
Caltech101               & 97.55 & 98.32 & \textbf{98.58} & 98.39 & 98.39 & 98.00  \\
StanfordCars             & 75.69 & 81.21 & 81.13 & 81.33 & \textbf{81.41} & 81.36  \\
SUN397                   & 80.99 & 82.58 & \textbf{82.81} & 82.54 & 82.67 & 82.33  \\
Food101                  & 90.49 & 91.09 & 91.15 & \textbf{91.18} & 91.12 & 91.08  \\
DTD                      & 80.09 & 84.95 & 84.61 & \textbf{85.76} & 85.53 & 83.22  \\
EuroSAT                  & 87.60 & 93.32 & 93.40 & \textbf{93.79} & 92.29 & 91.50  \\
Flowers102               & 96.96 & 98.10 & 98.86 & 98.96 & \textbf{98.77} & 98.67  \\
OxfordPets               & 95.06 & 95.11 & \textbf{95.80} & 95.48 & 95.27 & 94.90  \\
UCF101                   & 83.66 & 86.76 & \textbf{87.02} & 85.52 & 85.83 & 86.19  \\
FGVCAircraft             & 37.33 & 42.38 & \textbf{45.56} & 45.26 & 44.00 & 42.20  \\ 
\midrule
\rowcolor{gray!20}
Avg.                     & 81.98 & 84.69 & \textbf{85.15} & 85.11 & 84.84 & 84.28  \\
$\Delta$                       & \textcolor{V}{+0.00} & \textcolor{V}{+2.71} & \textcolor{V}{+3.17} & \textcolor{V}{+3.13} & \textcolor{V}{+2.86} & \textcolor{V}{+2.30}  \\
\bottomrule
\end{tabular}
    \caption{Ablation study on the impact of collaboration weight for base ($\omega_{b}$) of \texttt{DPC}. Benefiting from Weighting-Decoupling structure, weights for base or new can be different.}
    \label{tab8}
\end{table}
\begin{table}
\centering
\begin{tabular}{cc|ccc|c} 
\toprule
Method & weight & Base  & New   & H    & $\Delta$  \\ 
\midrule
MaPLe  &        & 83.52 & 73.31 & 78.08 &         \\
\textbf{+DPC}   & 0.2    & 85.07 & 73.31 & 78.75 & \textcolor{V}{+0.67}   \\
\rowcolor{gray!20}
\textbf{+DPC}   & 1.0    & \textbf{85.93} & \textbf{73.31} & \textbf{79.12} & \textcolor{V}{+1.04}   \\
\bottomrule
\end{tabular}
    \caption{Impact of collaboration weight for base ($\omega_{b}$) on \texttt{DPC} based on MaPLe \cite{khattak2023maple} backbone. Analysis of this phenomenon is exhibited in \cref{sec:B2}.}
    \label{tab9}
\end{table}
\begin{table}[th]
\setlength\tabcolsep{2.8pt}
\centering
\begin{tabular}{ccccccc} 
\toprule
\multirow{2.4}{*}{\textbf{Dataset}} & \multicolumn{6}{c}{\textbf{weight for new class ($\omega_{n}$)}}      \\ 
\cmidrule(lr){2-7}
                         & 0     & 0.01   & 0.02   & 0.05   & 0.1   & 0.2      \\ 
\midrule
ImageNet                 & \textbf{68.85} & 68.75 & 68.77 & 68.62 & 68.26 & 67.53  \\
Caltech101               & 94.65 & 94.98 & \textbf{95.09} & 94.98 & 94.87 & 94.76  \\
StanfordCars             & \textbf{70.14} & 69.79 & 69.42 & 68.61 & 66.92 & 63.04  \\
SUN397                   & \textbf{74.10} & 74.05 & 74.03 & 73.74 & 73.17 & 71.40  \\
Food101                  & 91.47 & 91.47 & 91.54 & 91.53 & \textbf{91.57} & 91.49  \\
DTD                      & 49.88 & 50.00 & \textbf{50.00} & 49.76 & 47.95 & 45.77  \\
EuroSAT                  & \textbf{51.62} & 51.41 & 51.08 & 49.31 & 46.05 & 39.79  \\
Flowers102               & \textbf{68.37} & 68.30 & 68.23 & 67.66 & 66.67 & 65.11  \\
OxfordPets               & 97.60 & 97.54 & 97.54 & \textbf{97.60} & 97.43 & 97.43  \\
UCF101                   & \textbf{66.31} & 65.66 & 65.33 & 64.09 & 63.66 & 62.52  \\
FGVCAircraft             & 24.24 & 23.94 & 24.06 & 24.12 & 24.25 & \textbf{25.85}  \\ 
\midrule
\rowcolor{gray!20}
Avg.                     & \textbf{68.84} & 68.72 & 68.64 & 68.18 & 67.35 & 65.88  \\
$\Delta$                        &\textcolor{V}{+0.00} & \textcolor{X}{-0.12} & \textcolor{X}{-0.20} & \textcolor{X}{-0.66} & \textcolor{X}{-1.49} & \textcolor{X}{-2.96}  \\
\bottomrule
\end{tabular}
    \caption{Ablation study on the impact of collaboration weight for new ($\omega_{n}$) of \texttt{DPC}. Benefiting from Weighting-Decoupling structure, weights for base or new can be different.}
    \label{tab10}
\end{table}
\begin{table}[th]
\setlength\tabcolsep{2.8pt}
\centering
\begin{tabular}{ccccccc} 
\toprule
\multirow{2.4}{*}{\textbf{Dataset}} & \multicolumn{6}{c}{\textbf{weight for target domain ($\omega_{n}$)}}   \\ 
\cmidrule(lr){2-7}
                         & 0     & 0.02  & 0.05  & 0.1   & 0.2   & 0.3    \\ 
\midrule
ImageNet-V2                 & \textbf{64.58} & 64.53 & 64.52 & 64.57 & 64.53 & 64.52  \\
ImageNet-S               & \textbf{48.89} & 48.83 & 48.83 & 48.77 & 48.72 & 48.61  \\
ImageNet-A             & \textbf{51.13} & 51.11 & 51.03 & 50.97 & 50.71 & 50.49  \\
ImageNet-R                   & \textbf{76.64} & 76.56 & 76.47 & 76.36 & 76.25 & 76.29  \\ 
\midrule
\rowcolor{gray!20}
Avg.                     & \textbf{60.31} & 60.26 & 60.21 & 60.17 & 60.05 & 59.98  \\
$\Delta$                         & \textcolor{V}{+0.00} & \textcolor{X}{-0.05} & \textcolor{X}{-0.10} & \textcolor{X}{-0.14} & \textcolor{X}{-0.26} & \textcolor{X}{-0.33}  \\
\bottomrule
\end{tabular}
    \caption{Ablation study on the impact of collaboration weight for target domain ($\omega_{n}$) of \texttt{DPC}. Benefiting from Weighting-Decoupling structure, weights for source or target can be different.}
    \label{tab10-2}
\end{table}

\noindent \textbf{Impact of Different Values.}
In \cref{tab8} and \cref{tab10}, we comprehensively compare the performance of various collaboration weights ($\omega_{b},\omega_{n}$) for base and new classes in \texttt{DPC} across 11 base-to-new tasks. For the base-class weight $\omega_{b}$, we observe that: \emph{(i)} Although the model achieves the best overall performance at $\omega_{b}=0.2$, this weight value is not necessarily representative of the peak performance for individual datasets. We attribute this to the diverse data distributions across different datasets. \emph{(ii)} When $\omega_{b}=1$, implying that the entire parallel prompt $\boldsymbol{P}^{\prime}$ is loaded for base class inference, the performance is still substantially better than the baseline. This corroborates that the Dynamic Hard Negative Optimizer in \texttt{DPC} effectually enhances the fitting of learnable prompts to the base classes.

In contrast, by observing the trend of the weight for new class $\omega_{n}$, we quantitatively verify the existence of the BNT problem, i.e. the model achieves maximum performance at $\omega_{n}=0$ (we add a 1$e$-6 term to avoid gradient propagation errors), and as the collaboration weight increases, gradually introducing the parallel prompt optimized on the base classes to the mixed prompt $\widetilde{\boldsymbol{P}}_{b}$, the performance of the model declines. We also acquire similar results in the ablation study of cross-domain transfer tasks in \cref{tab10-2}. This confirms that the optimization directions for base and new classes during fine-tuning are opposite, leading to interference between them. Nevertheless, benefiting from the Weighting-Decoupling architecture of \texttt{DPC}, the collaboration weights are variable across different tasks. Therefore, we directly set $\omega_{n} = 10^{-6}$ to retain generalization of backbones on new classes, successfully avoiding BNT problem.

\noindent \textbf{Special Phenomenon on MaPLe.}
For CoOp, PromptSRC, and PromptKD, we observe better performance at $\omega_{b}=0.2$ and $\omega_{n} = 10^{-6}$. However, for MaPLe, we discover that DPC achieves the best results at $\omega_{b}=1$, as exhibited in \cref{tab9}. Upon analysis, we consider that it may be due to the application of non-linear feature projection layer in MaPLe for generating visual prompts. Disrupting the linear consistency of latent feature channels between the visual and text prompt vectors (\S \ref{Sec4.4} in main text), this process leads to feature bias during the weighting of dual prompts.

\begin{figure}[th]
  \centering
  \includegraphics[width=0.97\linewidth]{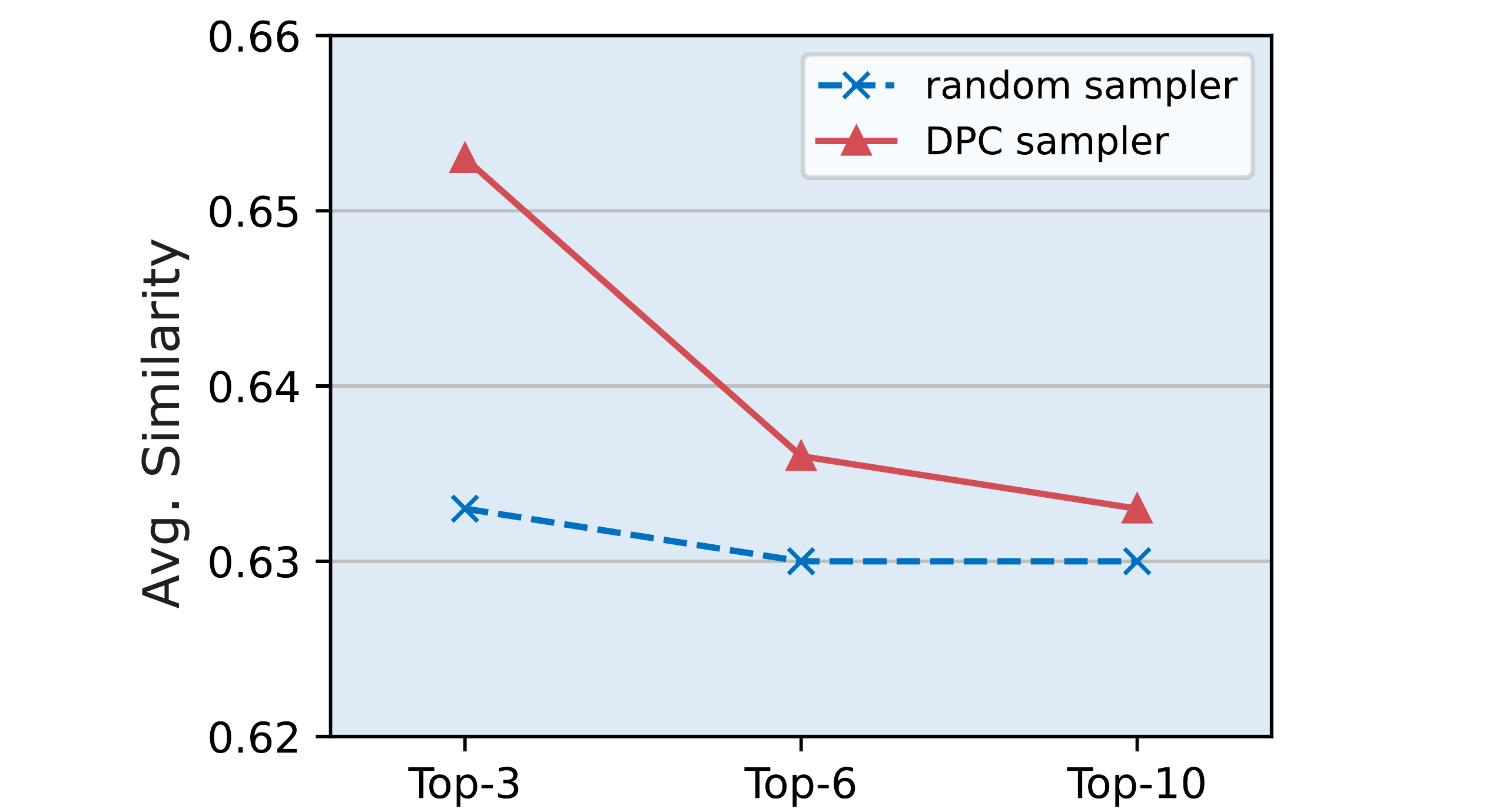}
  \caption{Cosine similarity between ground-truth and Top-$K$ results in entire Caltech101 \cite{fei2004caltech} dataset. We compare the similarity between random sampler in backbones and Negative Sampler in \texttt{DPC}. Higher score reveals stronger similarity.}
  \label{Figure 10}
\end{figure}
\begin{table}[t]
\centering
\setlength\tabcolsep{5pt}
\begin{tabular}{c|cc|cc} 
\toprule
\multirow{2.4}{*}{\textbf{Method}} & \multicolumn{2}{c|}{\textbf{branch}} & \multirow{2.4}{*}{\begin{tabular}[c]{@{}c@{}}\textbf{HM}\\\textbf{Acc.}\end{tabular}} & \multirow{2.4}{*}{$\Delta$}  \\ 
\cmidrule(lr){2-3}
                        & Text & Image               &                           &                                         \\ 
\midrule
PromptKD                &      & \checkmark                   & 83.59                     &                                         \\
\textbf{+DPC} (w/o img)           & \checkmark    &                     & 83.04                       & \textcolor{X}{-0.55}                                        \\
\textbf{+DPC} (w/ img)            & \checkmark    & \checkmark                   & \textbf{83.91}                     & \textcolor{V}{+0.32}                                   \\
\bottomrule
\end{tabular}
    \caption{Effect of freezing visual or text branches of \texttt{DPC} on base-to-new tasks, utilizing PromptKD \cite{li2024promptkd} as backbone model.}
    \label{tab11}
\end{table}

\subsection{Quantification of Negative Sampler}  \label{sec:B3}
In the Dynamic Hard Negative Optimizer module of \texttt{DPC}, the Negative Sampler is introduced for autonomously sampling hard negative objects (\S \ref{Sec3.3} in main text). To validate the effectiveness of this module, we quantify the discrepancy between the mini-batches sampled by \texttt{DPC} and the prompt tuning backbone using semantic similarity measurement. As demonstrated in \cref{Figure 10}, we apply a pre-trained bert-base-uncased \cite{devlin2018bert} model to calculate the average cosine similarity between ground-truth objects and other samples in the mini-batch obtained using either the Negative Sampler of \texttt{DPC} or the random sampling strategy of the backbone. Observations indicate that the samples obtained by \texttt{DPC} possess higher similarity, providing effective data-level gains for the Dynamic Hard Negative Optimizer.

\subsection{Effect of Visual or Text Branches on DPC}  \label{sec:B4}
To examine the impact of the \texttt{DPC} optimizer on the prompts in respective modality, we conduct ablation experiments on the visual and text branches based on \texttt{DPC-PK}. Specifically, after obtaining the tuned visual prompts through the PromptKD backbone, we attempt to freeze them and activate only the text branch during \texttt{DPC} fine-tuning, then compare this with the original \texttt{DPC} that activates both modality branches.

We notice in \cref{tab11} that freezing the visual branch results in the performance of \texttt{DPC} being even weaker than the backbone. We believe this is caused by the image-text contrastive loss of \texttt{DPC}, which enhances modality interaction and affects the feature channels of both branches simultaneously. Therefore, the operation that freezing single modality may lead to a deviation of text and image features. This indicates that the \texttt{DPC} optimizer simultaneously tunes both visual and text prompts, benefiting from the contrastive learning loss introduced by the Dynamic Hard Negative Optimizer.


\begin{table}
\centering
\setlength\tabcolsep{4pt}
\begin{tabular}{ccc|cc} 
\toprule
    & \begin{tabular}[c]{@{}c@{}}\textbf{Negative}\\\textbf{Sampling}\end{tabular} & \begin{tabular}[c]{@{}c@{}}\textbf{Hard Negative}\\\textbf{Optimizing }\end{tabular} & \begin{tabular}[c]{@{}c@{}}\textbf{HM}\\\textbf{Acc.}\end{tabular} & {$\Delta$}       \\ 
\midrule
(a) &                                                           & Cross Entropy                                                         & 74.84                                               &        \\
(b) & \checkmark                                                         & Cross Entropy                                                         & 75.06                                               & \textcolor{V}{+0.22}  \\
(c) & \checkmark                                                         & DPC Contrastive                                                           & \textbf{76.13}                                               & \textcolor{V}{+1.29}  \\
\bottomrule
\end{tabular}
    \caption{Additional ablation study on components in the Dynamic Hard Negative Optimizer. Experiments are conducted on the base-to-new tasks.}
    \label{tab14}
\end{table}

\begin{table}
\centering
\setlength\tabcolsep{2.5pt}

\begin{tabular}{cccccc} 
\toprule
\multirow{2.4}{*}{\textbf{Method}} & \multirow{2.4}{*}{\begin{tabular}[c]{@{}c@{}}\textbf{Learnable}\\\textbf{Params}\end{tabular}} & \multicolumn{3}{c}{\textbf{Memory Cost (MB)}} & \multirow{2.4}{*}{\begin{tabular}[c]{@{}c@{}}\textbf{Inference}\\\textbf{FPS}\end{tabular}}  \\ 
\cmidrule{3-5}
                                 &                                                                                              & ImgNet & Caltech & Cars                       &                                                                                            \\ 
\midrule
CoOp                             & 8K                                                                                           & 8126   & 1071    & 1813                       & 767.7                                                                                      \\
+DePT                            & (10+N/2) K                                                                                   & 8128   & 1195    & 1906                       & 773.2                                                                                      \\
\rowcolor{gray!20}
\textbf{+DPC}                    & 16K                                                                                          & 8390   & 1321    & 2067                       & 758.5                                                                                      \\
\bottomrule
\end{tabular}
    \caption{Computational cost comparison between CoOp backbone, DePT \cite{zhang2024dept} and our \texttt{DPC}. $N$ is the quantity of base classes.}
    \label{tab12}
\end{table}


\subsection{Ablation on Components in DHNO}  \label{sec:B5}
To demonstrate the necessity of each sub-module in the Dynamic Hard Negative Optimizer (DHNO) proposed in \S \ref{Sec3.3} of the main text, we conduct more ablation studies on the components of DHNO. The results are exhibited in \cref{tab14}. Since the Negative Sampler and Feature Filtering module are bound together in the process of reconstructing hard negatives, the Negative Sampling section in the table represents the combination of the two.

Compared with (a) CoOp backbone model, although (b) introducing only the Negative Sampler reveals a performance improvement, the gain is not distinct. We attribute this to the relatively weak effectiveness of the cross-entropy loss in the prompt learner backbones. Although the Negative Sampler effectively constructs mini-batches containing hard negatives, the standard cross-entropy loss, due to its lack of cross-modal interaction ability, fails to achieve deep alignment between visual and textual features. In contrast, significant enhancement in HM performance is observed in (c) introducing the symmetric image-text contrastive loss of \texttt{DPC}. The above results indicate a strong dependency among the Negative Sampler, Feature Filtering, and Hard Negative Optimizing components in DHNO. The combination of these 3 sub-modules leads to a remarkable improvement in base class performance.

\begin{table}
    \centering
    \setlength\tabcolsep{3pt}
\begin{tabular}{cc|ccc>{\columncolor{gray!20}}c} 
\toprule
\textbf{Datasets}                                                                &      & \begin{tabular}[c]{@{}c@{}}ProSRC\end{tabular} & +DePT & TCP   & \textbf{+DPC}   \\ 
\midrule
\multirow{3}{*}{\begin{tabular}[c]{@{}c@{}}Avg. over\\11 datasets\end{tabular}} & Base & 83.45                                                & 84.08 & 84.13 & \textbf{86.10}  \\
                                                                       & New  & 74.78                                                         & 75.03 & \textbf{75.36} & 74.78  \\
                                                                       & H    & 78.87                                                         & 79.29 & 79.51 & \textbf{80.04}  \\ 
\midrule
\multirow{3}{*}{ImageNet}                                              & Base & 77.28                                                         & 77.91 & 77.27 & \textbf{78.48}  \\
                                                                       & New  & 70.72                                                         & \textbf{70.77} & 69.87 & 70.72  \\
                                                                       & H    & 73.85                                                         & 74.17 & 73.38 & \textbf{74.40}  \\ 
\midrule
\multirow{3}{*}{Caltech101}                                            & Base & 97.93                                                         & 98.37 & 98.23 & \textbf{98.90}  \\
                                                                       & New  & 94.21                                                         & 94.14 & \textbf{94.67} & 94.21  \\
                                                                       & H    & 96.03                                                         & 96.21 & 96.42 & \textbf{96.50}  \\ 
\midrule
\multirow{3}{*}{OxfordPets}                                            & Base & 95.41                                                         & 94.83 & 94.67 & \textbf{96.13}  \\
                                                                       & New  & 97.30                                                         & 97.21 & 97.20 & \textbf{97.30}  \\
                                                                       & H    & 96.34                                                         & 96.00 & 95.92 & \textbf{96.71}  \\ 
\midrule
\multirow{3}{*}{StanfordCars}                                          & Base & 76.34                                                         & 78.26 & 80.80 & \textbf{82.28}  \\
                                                                       & New  & 74.98                                                         & 74.73 & 74.13 & \textbf{74.98}  \\
                                                                       & H    & 75.65                                                         & 76.46 & 77.32 & \textbf{78.46}  \\ 
\midrule
\multirow{3}{*}{Flowers102}                                            & Base & 97.06                                                         & 97.44 & \textbf{97.73} & 97.44  \\
                                                                       & New  & 73.19                                                         & 74.89 & \textbf{75.57} & 73.19  \\
                                                                       & H    & 83.45                                                         & 84.69 & \textbf{85.23} & 83.59  \\ 
\midrule
\multirow{3}{*}{Food101}                                               & Base & 90.83                                                         & 90.61 & 90.57 & \textbf{91.40}  \\
                                                                       & New  & 91.58                                                         & \textbf{91.63} & 91.37 & 91.58  \\
                                                                       & H    & 91.20                                                         & 91.12 & 90.97 & \textbf{91.49}  \\ 
\midrule
\multirow{3}{*}{Aircraft}                                              & Base & 39.20                                                         & 41.18 & 41.97 & \textbf{46.74}  \\
                                                                       & New  & 35.33                                                         & \textbf{35.63} & 34.43 & 35.33  \\
                                                                       & H    & 37.16                                                         & 38.20 & 37.83 & \textbf{40.24}  \\ 
\midrule
\multirow{3}{*}{SUN397}                                                & Base & 82.28                                                         & 82.60 & 82.63 & \textbf{83.63}  \\
                                                                       & New  & 78.08                                                         & \textbf{78.82} & 78.20 & 78.08  \\
                                                                       & H    & 80.13                                                         & 80.67 & 80.35 & \textbf{80.76}  \\ 
\midrule
\multirow{3}{*}{DTD}                                                   & Base & 83.45                                                         & 83.64 & 82.77 & \textbf{86.88}  \\
                                                                       & New  & 54.31                                                         & \textbf{59.18} & 58.07 & 54.31  \\
                                                                       & H    & 65.80                                                         & \textbf{69.32} & 68.25 & 66.84  \\ 
\midrule
\multirow{3}{*}{EuroSAT}                                               & Base & 92.84                                                         & 94.46 & 91.63 & \textbf{96.25}  \\
                                                                       & New  & 74.73                                                         & 71.01 & 74.73 & \textbf{74.73}  \\
                                                                       & H    & 82.80                                                         & 81.07 & 82.32 & \textbf{84.13}  \\ 
\midrule
\multirow{3}{*}{UCF101}                                                & Base & 85.28                                                         & 85.54 & 87.13 & \textbf{88.99}  \\
                                                                       & New  & 78.13                                                         & 77.29 & \textbf{80.77} & 78.13  \\
                                                                       & H    & 81.55                                                         & 81.20 & \textbf{83.83} & 83.21  \\
\bottomrule
\end{tabular}
    \caption{Detailed comparison between plug-and-play methods.}
    \label{tab15}
\end{table}

\subsection{Computational Cost}  \label{sec:B6}
\cref{tab12} summarizes the variations of learnable parameters, GPU memory overhead and inference time efficiency (evaluated by Frames Per Second, FPS) for the CoOp backbone, as well as two plug-and-play models, DePT and our \texttt{DPC}, across 3 example datasets. Due to the dual-prompt framework of \texttt{DPC}, the amount of learnable parameters in \texttt{DPC} is doubled relative to the initial model. However, profiting from the two-step fine-tuning strategy of \texttt{DPC}, the backbone prompt and parallel prompt are activated in separate stages, meaning that the computational overhead does not significantly increase. Experiments indicate that the memory cost of \texttt{DPC} slightly raises compared with the backbone ($\sim$ 0.25 GB), which we believe is mainly due to the increased computation required for the contrastive learning loss. As a PEFT method, the computational cost of introducing \texttt{DPC} to enhance prompt learners is completely acceptable.

\subsection{Detailed Comparison: Plug-and-Play}  \label{sec:B7}
In \cref{tab15}, we provide a more detailed supplement to the data presented in \cref{Figure 5} of the main text. Applying PromptSRC as the backbone model, we report the base-to-new performance of DePT and our \texttt{DPC} across 11 datasets, and introduce another plug-and-play model, TCP \cite{yao2024tcp}, for comparison. It is clear that \texttt{DPC} achieves superior base-class performance on most datasets, leading to the highest HM score among all baseline models.

\section{Limitation and Future Work} \label{sec:C}
Although our \texttt{DPC} effectively conquers the BNT problem in prompt tuning through prompt-level decoupling, we believe that this framework still has the room for optimization. Firstly, while we inherit the settings of the original backbone to obtain the tuned prompt, these configurations may not represent globally optimal points for generalization. How to adaptively acquire the top new-class performance through the backbone, thereby further leveraging the decoupled structure of \texttt{DPC}, remains a research-worthy question. Secondly, \texttt{DPC} demands learnable text prompts and image features (as well as optional visual prompts) for contrastive learning. For the research based on pure visual prompts (such as VPT \cite{jia2022vpt}) or feature extraction layers (such as CLIP-Adapter \cite{gao2024clipadapter}), it is challenging for \texttt{DPC} to integrally adapt as a plug-and-play approach. 

In future work, beyond the directions outlined in \cref{Sec5} of the main text, we will continue to explore strategies for enhancing the performance of base and new tasks, and investigate the feasibility of matching other forms of backbone models.

\end{document}